\documentclass[11pt,a4paper]{article}
\usepackage[table]{xcolor} 

\usepackage[hyperref]{emnlp2020}
\usepackage{times}
\usepackage{latexsym}

\usepackage{graphicx}
\usepackage{xspace}
\usepackage{paralist}
\usepackage{booktabs}
\usepackage{multirow}
\usepackage{amsmath}
\usepackage{amsfonts}
\usepackage{pifont}
\usepackage{amssymb}
\usepackage{mathrsfs}
\usepackage{makecell}
\usepackage{algorithm,algpseudocode}
\usepackage{mdwlist}

\usepackage{enumitem}
\usepackage{soul}

\usepackage{bm}
\usepackage{booktabs} 
\usepackage{multirow,multicol}
\usepackage{subfigure}
\usepackage{kotex}
\usepackage{mfirstuc,tabulary}
\usepackage{graphicx}
\usepackage{amsmath}
\usepackage{CJK}

\usepackage{microtype}

\newcommand{\mf}[1]{\multicolumn{2}{c}{ #1}}
\newcommand{\smf}[1]{\multicolumn{2}{c}{#1}}
\newcommand{\method}{mRASP\xspace}
\newcommand{\baseline}{Direct}

\newcommand{\dataset}{PC32\xspace}

\newcommand{\mml}{L}
\newcommand{\ml}{\mathcal{L}}
\newcommand{\mmd}{\mathcal{D}}
\newcommand{\mme}{\mathcal{E}}

\aclfinalcopy 

\newcommand{\xx}{\mathbf{x}}

\newcommand{\hide}[1]{}

\title{Pre-training Multilingual Neural Machine Translation by Leveraging Alignment Information}

\author{Zehui Lin$^{\dagger,\ddagger}$\thanks{\ \ Equal contribution. The work was done when the first author was an intern at ByteDance.} , 
Xiao Pan$^\dagger$\footnotemark[1] , 
Mingxuan Wang$^\dagger$, 
Xipeng Qiu$^\ddagger$,
Jiangtao Feng$^\dagger$,
Hao Zhou$^\dagger$,
Lei Li$^\dagger$ \\
  $^\dagger$ByteDance AI Lab \\
  \textit{\{panxiao.94,wangmingxuan.89,zhouhao.nlp,fengjiangtao,lileilab\}@bytedance.com} \\
  $^\ddagger$School of Computer Science, Fudan University, Shanghai, China \\
  \textit{\{linzh18,xpqiu\}@fudan.edu.cn} \\
  }

\date{}

\begin{document}

\maketitle

\begin{abstract}

We investigate the following question for machine translation (MT):
can we develop a single universal MT model to serve as the common seed and obtain derivative and improved models on arbitrary language pairs?
We propose \method, an approach to pre-train a universal multilingual neural machine translation model.
Our key idea in \method is its novel technique of random aligned substitution, which brings words and phrases with similar meanings across multiple languages closer in the representation space. 
We pre-train a \method model on 32 language pairs jointly with only public datasets. 
The model is then fine-tuned on downstream language pairs to obtain specialized MT models.
We carry out extensive experiments on 42 translation directions across a diverse settings, including low, medium, rich resource, and as well as transferring to exotic language pairs.  
Experimental results demonstrate that \method
 achieves significant performance improvement
 compared to directly training on those target pairs. 
It is the first time to verify that multilingual MT can be utilized to improve rich resource MT.
We expand the notion of "zero-shot translation" in multilingual NMT for the first time to "exotic translation" and categorize it into four scenarios.
Surprisingly, \method is even able to improve the translation quality on exotic languages that never occur in the pre-training corpus. 
Code, data, and pre-trained models are available at \url{https://github.com/linzehui/mRASP}.

\end{abstract}

\section{Introduction}
\label{sec:intro}

Pre-trained language models such as BERT have been highly effective for NLP tasks~\cite{DBLP:conf/naacl/PetersNIGCLZ18,DBLP:conf/naacl/DevlinCLT19,radford2019language,DBLP:conf/nips/ConneauL19,Liu2019RoBERTaAR,DBLP:conf/nips/YangDYCSL19}.
Pre-training and fine-tuning has been a successful paradigm. 
It is intriguing to discover a ``BERT'' equivalent -- a pre-trained model -- for machine translation. 
In this paper, we study the following question: 
can we develop a single universal MT model and derive specialized models by fine-tuning on an arbitrary pair of languages?

\hide{
Pre-training language models (LMs), 
such as ELMo~, BERT~\cite{DBLP:conf/naacl/DevlinCLT19}, 
GPT~\cite{radford2019language}, Cross-lingual Language Model (XLM) ~\cite{DBLP:conf/nips/ConneauL19},
RoBERTa~\cite{Liu2019RoBERTaAR} and XLNet~\cite{DBLP:conf/nips/YangDYCSL19} have  been highly successful in machine learning and natural language processing communities. 
Typical pre-trained LMs employ raw sentences (which is always very cheap) and cleverly design objective to obtain knowledge from them. 
}

While pre-training techniques are working very well for NLP task, there are still several limitations for  machine translation tasks. 
First, pre-trained language models such as BERT are not easy to directly fine-tune unless using some sophisticated techniques~\citep{Yang2019TowardsMT}.
Second, there is a discrepancy between existing pre-training objective and down-stream ones in MT. 
Existing pre-training approaches such as MASS~\cite{DBLP:conf/icml/SongTQLL19} and mBART~\cite{DBLP:journals/corr/abs-2001-08210} rely on auto-encoding objectives to pre-train the models, which are different from translation. 
Therefore, their fine-tuned MT models still do not achieve adequate improvement. 
Third, existing MT pre-training approaches focus on using multilingual models to improve MT for low resource or medium resource languages. 
There has not been one pre-trained MT model that can improve for any pairs of languages, even for rich resource settings such as English-French. 

\hide{
Despite its success, at least two challenges still exist when directly using pre-trained LM in neural machine translation~(NMT).  First, the objective of pre-training is different from the downstream NMT tasks. Some recent attempts propose sequence-to-sequence pre-training methods to alleviate this problem. MASS uses an auto-encoder to pre-train the sequence model~\cite{DBLP:conf/icml/SongTQLL19}. mBART extends the monolingual pre-training to a large scale multilingual approach~\cite{DBLP:journals/corr/abs-2001-08210}. However, these approaches' objective is a monolingual auto-encoder, which is still different from the translation. 
Second, most previous pre-training approaches focus on improving low resource or medium resource languages.   MASS only conducts experiments on unsupervised and low-resource languages. mBART pre-trained with more data than MASS shows impressive improvements on low-resource NMT, but slight degradation on rich-resource language pairs, such as WMT14 En-Fr and En-De.
}


In this paper, we propose multilingual Random Aligned Substitution Pre-training (\method), a method to pre-train a MT model for many languages, which can be used as a common initial model to fine-tune on arbitrary language pairs. 
\method will then improve the translation performance, comparing to the  MT models directly trained on downstream parallel data. In our method, we ensure that the pre-training on many languages and the down-stream fine-tuning share the same model architecture and training objective. 
Therefore, this approach lead to large translation performance gain. 
Consider that many languages differ lexically but are closely related at the semantic level, we start by training a large-scale multilingual NMT model across different translation directions, then fine-tuning the model in a specific direction.
Further, to close the representation gap across different languages and make full use of multilingual knowledge, we explicitly introduce additional loss based on random aligned substitution of the words in the source and target sentences. 
Substituted sentences are trained jointly with the same translation loss as the original multilingual parallel corpus.
In this way, the model is able to bridge closer the representation space across different languages.

\begin{figure*}[htbp]
    \centering
    \includegraphics[width=1.0\textwidth]{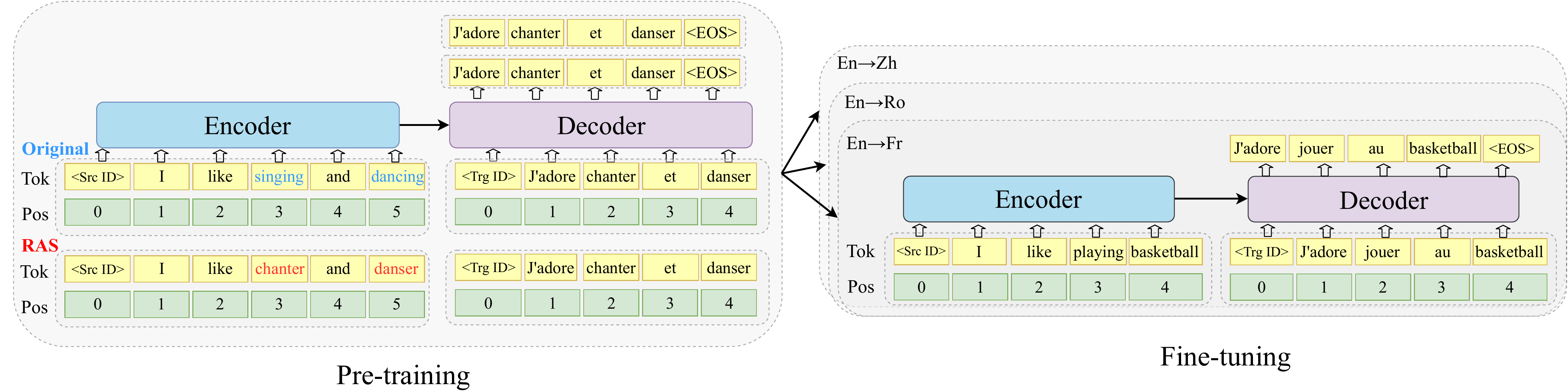}
    \caption{The proposed \method method. ``Tok'' denotes token embedding while ``Pos'' denotes position embedding. During the pre-training phase, parallel sentence pairs in many languages are trained using translation loss, together with their substituted ones. We randomly substitute words with the same meanings in the source and target sides. 
    During the fine-tuning phase, we further train the model on the downstream language pairs to obtain specialized MT models.}
    \label{fig:framework}
\end{figure*}

We carry out extensive experiments in different scenarios, including translation tasks with different dataset scales, as well as exotic translation tasks.
 For extremely low resource ($<$100k), \method obtains gains up to +22 BLEU points compared to directly trained models on the downstream language pairs.
 \method obtains consistent performance gains as the size of datasets increases. 
 Remarkably, even for rich resource ($>$10M, e.g. English-French), \method still achieves big improvements. 

We divide "exotic translation" into four categories with respect to the source and target side.
\begin{itemize}
    \setlength{\itemsep}{0pt}
    \setlength{\parsep}{0pt}
    \setlength{\parskip}{0pt}
\item \textbf{Exotic Pair} Both source and target languages are individually pre-trained while they have not been seen as bilingual pairs.
\item \textbf{Exotic Source} Only target language is pre-trained, but source language is not.
\item \textbf{Exotic Target} Only source language is pre-trained, but the target language is not.
\item \textbf{Exotic Full} Neither source nor target language is pre-trained. 
\end{itemize}

 Surprisingly, even when \method is fine-tuned on "exotic full" language pair, the resulting MT model is still much better than the directly trained ones (+3.3 to +14.1 BLEU). 
We finally conduct extensive analytic experiments to examine the contributing factors inside the \method method for the performance gains.

We highlight our contributions as follows:
\begin{itemize}
    \item We propose \method, an effective  pre-training method that can be utilized to fine-tune on any language pairs in NMT. 
    It is very efficient in the use of parallel data in multiple languages. 
    While other pre-trained language models are obtained through hundreds of billions of monolingual or cross-lingual sentences, \method only introduces several hundred million bilingual pairs. We suggest that the consistent objectives of pre-training and fine-tuning lead to better model performance.
    \item We explicitly introduce a random aligned substitution technique into the pre-training strategy, and find that such a technique can bridge the semantic space between different languages and thus improve the final translation performance.
    \item We conduct extensive experiments 42 translation directions across different scenarios, demonstrating that \method can significantly boost the performance on various translation tasks. \method achieves 14.1 BLEU with only 12k  pairs of Dutch and Portuguese sentences even though neither appears in the pre-training data. 
    \method also achieves 44.3 BLEU on WMT14 English-French translation. 
    Note that our pre-trained model only use parallel corpus in 32 languages, unlike other methods that also use much more monolingual raw corpus. 
\end{itemize}

\section{Methodology}
\label{sec:approach}

In this section, we introduce our proposed \method and the training details.


\subsection{\method}

\paragraph{Architecture}
We adopt a standard Transformer-large architecture~\cite{DBLP:conf/nips/VaswaniSPUJGKP17} with 6-layer encoder and 6-layer decoder. 
The model dimension is 1,024 on 16 heads. 
We replace ReLU with GeLU \cite{DBLP:journals/corr/HendrycksG16} as activation function on feed forward network. 
We also use learned positional embeddings.

\paragraph{Methodology}
A multilingual neural machine translation model learns a many-to-many mapping function $f$ to translate from one language to another. More formally, define  $\mml=\left\{\mml_{1}, \ldots, \mml_{M}\right\}$ where $\mml$ is a collection of languages involving in the pre-training phase. 
$\mmd_{i,j}$ denotes a parallel dataset of $(\mml_{i}, \mml_j)$, and $\mme$ denotes the set of parallel datasets $\{\mmd\}_{i=1}^{i=N}$, where $N$ the numbers of the bilingual pair. The training loss is then defined as:
\begin{equation}
    \ml^{pre} = \sum_{i,j \in \mme} \mathbb{E}_{(\xx^i,\xx^j)\sim \mmd_{i,j}} [-\log P_\theta(\xx^i|C(\xx^j))].
\label{eq:multi}
\end{equation}
where $\xx^i$ represents a sentence in language $\mml_i$, and $\theta$ is the parameter of \method, and $C(\xx^i)$ is our proposed alignment function, which randomly replaces the words in $\xx^i$ with a different language. 
In the pre-training phase, the model jointly learns all the translation pairs.

\paragraph{Language Indicator}
Inspired by \cite{DBLP:journals/tacl/JohnsonSLKWCTVW17, ha2016toward},  to distinguish from different translation pairs, we simply add two artificial language tokens to indicate languages at the source and target side. For instance, the following En$\rightarrow$Fr sentence  
``\texttt{How are you? -> Comment vas tu?} ''
is transformed to 
``\texttt{<en>  How are you? -> <fr> Comment vas tu?}''

\paragraph{Multilingual Pre-training via RAS}

Recent work proves that cross-lingual language model pre-training could be a more effective way to representation learning ~\cite{DBLP:conf/nips/ConneauL19,huang2019unicoder}.
However, the cross-lingual information is mostly obtained from shared subword vocabulary during pre-training, which is limited in several aspects:
\begin{itemize}
    \setlength{\itemsep}{0pt}
    \setlength{\parsep}{0pt}
    \setlength{\parskip}{0pt}
    \item The vocabulary sharing space is sparse in most cases. Especially for dissimilar language pairs, such as English and Hindi, they share a fully different morphology. 
    \item The same subword across different languages may not share the same semantic meanings. 
    \item The parameter sharing approach lacks explicit supervision to guild the word with the same meaning from different languages shares the same semantic space.
\end{itemize}

Inspired by constructive learning, we propose to bridge the semantic gap among different languages through \textbf{Random Aligned Substitution (RAS)}. 
Given a parallel sentence $(\xx^i, \xx^j)$,  we randomly replace a source word in $\xx^i_t$ to a different random language $L_k$, where $t$ is the word index. We adopt an unsupervised word alignment method MUSE\cite{conneau2017word}, which can  translate $\xx^i_t$ to $d_{i,k}(\xx^i_t)$ in language $L_k$, where $d_{i,k}(\cdot)$ is the dictionary translating function. 
With the dictionary replacement, the original bilingual pair will construct a code-switched sentence pair $(C(\xx^i), \xx^j)$.  
As the benefits of  random sampling, the translation set  $\{d_{i,k}(\xx^i_t)\}_{k=1}^{k=M}$ potentially appears in the same context. 
Since the word representation depends on the context, the word with similar meaning across different languages can share a similar representation. Figure \ref{fig:framework} shows our alignment methodology.




\begin{table*}[htb]
\begin{center}
\begin{tabular}{rcccccccccccc}
\toprule
&\multicolumn{7}{c}{\textbf{Extremely Low Resource ($<$100k)}} \\
\midrule

Lang-Pairs& 
\mf{En-Be} &
\mf{En-My} &
\mf{En-Af} &
\mf{En-Eo} & 
Avg

\\

%

Size &
\smf{20K}  & 
\smf{29k}  & 
\smf{41K} & 
\smf{67K} & 

\\

Direction & 
 $\rightarrow$ &$\leftarrow$ &

 $\rightarrow$ &$\leftarrow$ &

 $\rightarrow$ &$\leftarrow$ &

 $\rightarrow$ &$\leftarrow$ \\

\midrule
\baseline &
6.9 & 4.4 & 
12.0 & 3.6 &  
8.9 & 8.0 & 
5.9 & 6.0 & 
7.0
\\

 \method &
\bf 24.7 & \bf 34.1 &
\bf 28.4 & \bf 26.1 &
\bf 30.4 & \bf 26.3 &
\bf 30.8 & \bf 35.1 &
29.5

\\

 $\Delta$ &
 +17.8 & +29.7 &
 +16.4 & +22.5  &
 +21.5 & +18.3 &
 +24.9 & +29.1 &
 \textbf{+22.5}

\\

\midrule
\midrule

&\multicolumn{7}{c}{\textbf{Low Resource (100k$\sim$1m)}} \\
\midrule

Lang-Pairs &
\mf{En-He} &
\mf{En-Tr} &
\mf{En-Ro} &
\mf{En-Cs} &
Avg
\\

Size &
\smf{335K} & 
\smf{388K} & 
\smf{600K} & 
\smf{978K}
\\

 Direction & 
 $\rightarrow$ &$\leftarrow$ &

 $\rightarrow$ &$\leftarrow$ &

 $\rightarrow$ &$\leftarrow$ &

 $\rightarrow$ &$\leftarrow$ \\

\midrule
 \baseline &
28.8 & 40.6 & 
14.0 & 30.1 & 
30.5 & 29.2 & 
20.4 & 26.2 & 
27.5
\\

 \method &
\bf 32.4 & \bf 44.6 &  
\bf 21.0 & \bf 33.3 & 
\bf 39.0 & \bf 37.4 & 
\bf 23.2 & \bf 29.8 & 
33.1

\\

$\Delta$ &
+3.8 & +4.3 &
+5.4 & +3.7  &
+8.5 & +8.2 &
+5.3 & +5.6 &
\textbf{+5.6}

\\

\midrule
\midrule
&\multicolumn{7}{c}{\textbf{Medium Resource (1m$\sim$10m)}} \\
\midrule

Lang-Pairs & 
\mf{En-Ar} &
\mf{En-Et} &
\mf{En-Bg} &
\mf{En-De} &
Avg
\\

Size &
\smf{1.2M}  & 
\smf{2.3M} & 
\smf{3.1M} & 
\smf{4.5M}  

\\

Direction & 
 $\rightarrow$ &$\leftarrow$ &

 $\rightarrow$ &$\leftarrow$ &

 $\rightarrow$ &$\leftarrow$ &

 $\rightarrow$ &$\leftarrow$ \\

\midrule
 \baseline &
18.2 & 35.0 & 
24.1 & 26.7 & 
38.3 & 40.5 & 
29.3 & 30.8 &  
30.4
\\

 \method &
\bf 20.0 & \bf 38.7 & 
\bf 26.3 & \bf 33.2 & 
\bf 40.1 & \bf 44.3 & 
\bf 30.3 & \bf 34.4 & 
33.4
\\

$\Delta$ &
+1.8 & +3.7 &
+2.2 & +6.5 &
+1.8 & +3.8 &
+1.0 & +3.6 &
\textbf{+3.1}

\\

\bottomrule
\end{tabular}
\caption{Fine-tuning performance on \emph{extremely low} / \emph{low} / \emph{medium} resource machine translation settings. The numbers in parentheses indicate the size of parallel corpus for fine-tuning.   
Pre-training with \method and then fine-tuning on downstream MT tasks consistently improves over MT models directly trained on bilingual parallel corpus. }
\label{tab:LowR}
\end{center}
\end{table*}

\subsection{Pre-training Data}

We collect 32 English-centric language pairs, resulting in 64 directed translation pairs in total. English is served as an anchor language bridging all other languages. 
The parallel corpus are from various sources: \textit{ted}\footnote{Compiled by \citet{DBLP:conf/naacl/QiSFPN18}. For simplicity, we deleted zh-tw and zh (which is actually Cantonese), and merged fr-ca with fr, pt-br with pt.}, \textit{wmt}\footnote{\url{http://www.statmt.org}}, \textit{europarl}\footnote{\url{http://opus.nlpl.eu/Europarl-v8.php}}, \textit{paracrawl}\footnote{\url{https://paracrawl.eu/}}, \textit{open-subtitles}\footnote{\url{http://opus.nlpl.eu/OpenSubtitles-v2018.php}}, \textit{qed}\footnote{\url{http://opus.nlpl.eu/QED-v2.0a.php}}. 
We refer to our pre-training data as \textbf{\dataset}(Parallel Corpus 32). 
\textbf{\dataset} contains a total size of 197M pairs of sentences.
Detailed descriptions and summary for the datasets can be found in Appendix.

For RAS, we utilize ground-truth En-X bilingual dictionaries\footnote{\url{https://github.com/facebookresearch/MUSE}}, where X denotes languages involved in \dataset. Since not all languages in \dataset have ground-truth dictionaries, we only use available dictionaries.


\subsection{Pre-training Details}

We use learned joint vocabulary. We learn shared BPE \cite{sennrich2015neural} merge operations (with 32k merge ops) across all the training data and added monolingual data as a supplement (limit to 1M sentences).
We do over-sampling in learning BPE to balance the vocabulary size of languages, whose resources are drastically different in size. We over-sampled the corpus of each language based on the volume of the largest language corpus. We keep tokens occurring more than 20, which results in a subword vocabulary of 64,808 tokens.

In pre-training phase, we train our model with the full pairs of the parallel corpus. Following the training setting in Transformer, we use Adam optimizer with $\epsilon=1e-8,\beta_2 = 0.98$. A warm-up and linear decay scheduling with a warm-up step of 4000 is used. We pre-train the model for a total of 150000 steps. 

For RAS, we use the top 1000 words in dictionaries and only substitute words in source sentences. Each word is replaced with a probability of 30$\%$ according to the En-X bilingual dictionaries. To address polysemy, we randomly select one substitution from all candidates.

\section{Experiments}
\label{sec:exps}

This section shows that \method obtains consistent performance gains in different scenarios. We also compare our method with existing pre-training methods and outperforms the baselines on En$\rightarrow$Ro dataset.
The performance further boosts by combining back-translation\cite{DBLP:conf/acl/SennrichHB16} technique.
Otherwise stated, for all experiments, we use the pre-trained model as initialization and fine-tune with the downstream target parallel corpus.
\subsection{Experiment Settings}

\begin{table}[ht!]
\centering
\begin{tabular}{rccccc}
\toprule
 Lang-Pairs &  En$\rightarrow$De&  Zh$\rightarrow$En &  En$\rightarrow$Fr \\ 
Size &4.5M& 20M & 40M \\
\midrule
\baseline &29.3 & 24.1  & 43.2  \\
CTNMT\footnotemark~\shortcite{Yang2019TowardsMT} & 30.1 &  - & 42.3   \\
mBART~\shortcite{DBLP:journals/corr/abs-2001-08210} &-&  - & 41.0   \\
XLM~\shortcite{DBLP:conf/nips/ConneauL19}   & 28.8 & - & - \\
MASS~\shortcite{DBLP:conf/icml/SongTQLL19}   & 28.9 & - & - \\
mBERT~\shortcite{DBLP:conf/naacl/DevlinCLT19}   & 28.6 & - &  -\\
\midrule
\method &\bf 30.3 & \bf 24.7 &  \textbf{44.3}   \\
\bottomrule
\end{tabular}
\caption{Fine-tuning performance for popular \emph{medium} and \emph{rich} resource MT tasks. For fair comparison, we report detokenized BLEU on WMT newstest18 for Zh$\rightarrow$En and tokenized BLEU on WMT newstest14 for En$\rightarrow$Fr and En$\rightarrow$De.
Notice unlike previous methods (except CTNMT)  which do not improve in the rich resource settings, \method  is again able to consistently improve the downstream MT performance. 
It is the first time to verify that low-resource language pairs can be utilized to improve rich resource MT.
}
\label{tab:RichR}
\end{table}
\footnotetext{CTNMT only reports the Transformer-base setting.}

\begin{table*}[ht]
\begin{center}
\begin{tabular}{rcccccccccccc}
\toprule
& \multicolumn{4}{c}{\textbf{Exotic Pair}} & \multicolumn{4}{c}{\textbf{Exotic Full}} \\
\midrule

Lang-Pairs& 
\mf{Fr-Zh} &
\mf{De-Fr} &
\mf{Nl-Pt} &
\mf{Da-El} & 

\\

%

Size &
\smf{20K}  & 
\smf{9M}  & 
\smf{12K} & 
\smf{1.2M} & 

\\

Direction & 
 $\rightarrow$ &$\leftarrow$ &

 $\rightarrow$ &$\leftarrow$ &

 $\rightarrow$ &$\leftarrow$ &

 $\rightarrow$ &$\leftarrow$ \\

\midrule
\baseline &
0.7 & 3.0 & 
23.5 & 21.2 &  
0.0 & 0.0 & 
14.1 & 16.9 

\\

\method &
\bf 25.8 & \bf 26.7 &
 \bf 29.9 & \bf 23.4 &
\bf 14.1 & \bf 13.2 &
\bf 17.6 &\bf 19.9

\\
\midrule
\midrule

& \multicolumn{8}{c}{\textbf{Exotic Source/Target}} \\
\midrule

Lang-Pairs &
\mf{En-Mr} &
\mf{En-Gl} &
\mf{En-Eu} &
\mf{En-Sl}
\\

Size &
\smf{11K} & 
\smf{200K} & 
\smf{726K} & 
\smf{2M}
\\

 Direction & 
 $\rightarrow$ &$\leftarrow$ &

 $\rightarrow$ &$\leftarrow$ &

 $\rightarrow$ &$\leftarrow$ &

 $\rightarrow$ &$\leftarrow$ \\

\midrule
\baseline &
6.4 & 6.8 & 
8.9 & 12.8 & 
7.1 & 10.9 & 
24.2 & 28.2  
 
\\

\method &
\bf 22.7 & \bf 22.9 & 
\bf 32.1 & \bf 38.1 & 
\bf 19.1 & \bf 28.4 & 
\bf 27.6 & \bf 29.5 

\\

\bottomrule
\end{tabular}
\caption{Fine-tuning MT performance on exotic language corpus. For two the translation direction A$\rightarrow$B, 
\emph{exotic pair}: A and B occur in the pre-training corpus but no pairs of sentences of (A,B) occur; \emph{exotic full}: no sentences in either A nor B occur in the pre-training; \emph{exotic source}: sentences from the target side B occur in the pre-training but not the source side A; \emph{exotic target}: sentences from the source side A occur in the pre-training but not the target side B.
Notice that pre-training with \method and fine-tuning on those exotic languages consistently obtains significant improvements MT performance in each category.
}
\label{tab:Unseen}
\end{center}
\end{table*}
\begin{table}[ht]
\centering
\scalebox{0.88}{
\begin{tabular}{l|cccc}
\toprule
Model & En$\rightarrow$Ro &  Ro$\rightarrow$En & Ro$\rightarrow$En +BT  \\ 
\midrule
\baseline  & 34.3 & 34.0 & 36.8 \\
\midrule
XLM~\shortcite{DBLP:conf/nips/ConneauL19}   & - & 35.6 & 38.5 \\
MASS~\shortcite{DBLP:conf/icml/SongTQLL19}   & - & - & \bf 39.1\\
BART~\shortcite{DBLP:conf/acl/LewisLGGMLSZ20}   & - & - & 38.0\\
XLM-R~\shortcite{DBLP:conf/acl/ConneauKGCWGGOZ20}  & 35.6 & 35.8 & - \\
mBART~\shortcite{DBLP:journals/corr/abs-2001-08210} & \bf 37.7 & \bf 37.8 & 38.8 \\

\midrule
\method  &  37.6 & 36.9 & 38.9 \\

\bottomrule
\end{tabular}
}
\caption{Comparison with previous Pre-training models on WMT16 En-Ro. Following \cite{DBLP:journals/corr/abs-2001-08210}, We report detokenized BLEU. We reaches comparable results on both  En$\rightarrow$Ro and Ro$\rightarrow$En. By combining back translation, the performance further boost for 2 BLEU points on Ro$\rightarrow$En. We remove diacritics for Romanian corpus during training and inference and report En$\rightarrow$Ro BLEU score under this condition.
}
\label{tab:vs_other}
\vspace{-10pt}
\end{table}

\paragraph{Datasets}
We collect 14 pairs of parallel corpus to simulate different scenarios. Most of the En-X parallel datasets are from the pre-training phase to avoid introducing new information. Most pairs for fine-tuning are from previous years of WMT and IWSLT. Specifically, we use WMT14 for En-De and En-Fr, WMT16 for En-Ro. For pairs like Nl(Dutch)-Pt(Portuguese) that are not available in WMT or IWSLT, we use news-commentary instead. 
For a detailed description, please refer to the Appendix.

Based on the volume of parallel bi-texts, we divide the datasets into four categories: extremely low resource ($<$100K), low resource($>$100k and $<$1M), medium resource ($>$1M and $<$10M), and rich resource ($>$10M).

For back translation, we include 2014-2018 newscrawl for the target side, En. The total size of the monolingual data is 3M.

\paragraph{Baseline}

To better quantify the effectiveness of the proposed pre-training models, we also build two baselines.

\textbf{\method w/o RAS}. To measure the effect of alignment information, we also pre-train a model on the same \dataset. We do not include alignment information on this pre-training model.

\textbf{\baseline}.
We also train randomly initialized models directly on downstream bilingual parallel corpus as a comparison with pre-training models.

\paragraph{Fine-tuning}
We fine-tune our obtained \method model on the target language pairs. 
We apply a dropout rate of 0.3 for all pairs except for rich resource such as En-Zh and En-Fr with 0.1. We carefully tune the model, setting different learning rates and learning scheduler warm-up steps for different data scale. For inference, we use beam-search with beam size 5 for all directions.
For most cases, We measure case-sensitive tokenized BLEU. We also report de-tokenized BLEU with SacreBLEU \cite{DBLP:conf/wmt/Post18} for a fair comparison with previous works.

\subsection{Main Results}

We first conduct experiments on the (extremely) low-resource and medium-resource datasets, where multilingual translation usually obtains significant improvements. As illustrated in Table \ref{tab:LowR}, we obtain significant gains in all datasets. 
For extremely low resources setting such as En-Be (Belarusian) where the amount of datasets cannot train an NMT model properly, utilizing the pre-training model boosts performance.

We also obtain consistent improvements in low and medium resource datasets. Not surprisingly, We observe that with the scale of the dataset increasing, the gap between the randomly initialized baseline and pre-training model is becoming closer. It is worth noting that, for En$\rightarrow$De benchmark, we obtain 1.0 BLEU points gains\footnote{We report results of En$\rightarrow$De on newstest14. The baseline result is reported in \citet{DBLP:conf/wmt/OttEGA18}}. Extra experiment results on public testsets are provided in Table~\ref{tab:LowRpub}.

To verify \method can further boost performance on rich resource datasets, we also conduct experiments on En$\rightarrow$Zh and En$\rightarrow$Fr. We compare our results with two strong baselines reported by \citet{DBLP:conf/wmt/OttEGA18, DBLP:conf/wmt/LiLXLLLWZXWFCLL19}. As shown in Table \ref{tab:RichR}, surprisingly, when large parallel datasets are provided, it still benefits from pre-training models. In En$\rightarrow$Fr, we obtain 1.1 BLEU points gains.

\paragraph{Comparing to other Pre-training Approaches}
We compare our \method to recently proposed multilingual pre-training models. Following \citet{DBLP:journals/corr/abs-2001-08210}, we conduct experiments on En-Ro, the only pairs with established results. To make a fair comparison, we report de-tokenized BLEU.

As illustrated in Table \ref{tab:vs_other} , Our model reaches comparable performance on both En$\rightarrow$Ro and Ro$\rightarrow$En. We also combine Back Translation \cite{DBLP:conf/acl/SennrichHB16} with  \method, observing performance boost up to 2 BLEU points, suggesting \method is complementary to BT. It should be noted that the competitors introduce much more pre-training data.

mBART contucted experiments on extensive language pairs. To illustrate the superiority of \method, we also compare our results with mBART. We use the same test sets as mBART. As illustrated in Table \ref{tab:cmp-mBART}, mRASP outperforms mBART for most of language pairs by a large margin. Note that while mBART underperforms baseline for benchmarks En-De and En-Fr, \method obtains 4.3 and 2.9 BLEU gains compared to baseline.

\begin{table}[ht]
\begin{center}
\scalebox{0.9}{
\begin{tabular}{lcccccc}
\toprule

Lang-Pairs& 

\mf{En-Gu} & 
\mf{En-Kk} &
\mf{En-Tr}

\\

Source & 
\mf{WMT19}   & 
\mf{WMT19}   & 
\mf{WMT17}  

\\



 Direction & 
 $\rightarrow$ &$\leftarrow$ &

 $\rightarrow$ &$\leftarrow$ &

 $\rightarrow$ &$\leftarrow$ \\

\midrule
 \baseline &
0.0 & 0.0 & 
0.2 & 0.8 &  
9.5 & 12.2 

\\

 mBART &
 0.1 & 0.3  &
  2.5 &  7.4 &
 17.8 &  22.5

\\

 \method&
\textbf{3.2} & \textbf{0.6} &
\textbf{8.2} & \textbf{12.3} &
\textbf{20.0} & \textbf{23.4}

\\

\midrule
\midrule

Lang-Pairs& 

\mf{En-Et} & 
\mf{En-Fi} &
\mf{En-Lv}

\\

Source & 
\mf{WMT18}   & 
\mf{WMT17}   & 
\mf{WMT17}  

\\

 Direction & 
 $\rightarrow$ &$\leftarrow$ &

 $\rightarrow$ &$\leftarrow$ &

 $\rightarrow$ &$\leftarrow$ \\

\midrule

 \baseline &
17.9 & 22.6 & 
20.2 & 21.8 &  
12.9 & 15.6 

\\

mBART &
 21.4 & 27.8  &
  22.4 &  28.5 &
15.9 &  19.3

\\

 \method &
\textbf{21.9} & \textbf{28.2} &
 \textbf{24.0} &  28.0 &
 \textbf{21.6} & \textbf{24.4}

\\

\midrule
\midrule

Lang-Pairs & 

\mf{En-Cs} & 
\mf{En-De} &
\mf{En-Fr}

\\

Source & 
\mf{WMT19}   & 
\mf{WMT19}   & 
\mf{WMT14}  

\\

 Direction & 
 \mf{$\rightarrow$} & 

 \mf{$\rightarrow$} &

 \mf{$\rightarrow$} 
 
\\

\midrule

 \baseline &
\mf{16.5} & 
\mf{30.9} &  
\mf{41.4}   

\\

mBART &
\mf{18.0}  &
\mf{30.5} & 
\mf{41.0}

\\

\method &
\mf{\textbf{19.9}} & 
\mf{\textbf{35.2}} &  
\mf{\textbf{44.3}}

\\

\bottomrule
\end{tabular}
}
\caption{Comprehensive comparison with mBART. \method  outperforms mBART on MT for all but two language pairs.}
\label{tab:cmp-mBART}
\end{center}
\end{table}


\subsection{Generalization to Exotic Translation}

To illustrate the generalization of \method, we also conduct experiments on exotic translation directions, which is not included in our pre-training phase. 
For each category, we select language pairs of different scales.

The results are shown in Table \ref{tab:Unseen}. As is shown, \method obtains significant gains for each category for different scales of datasets, indicating that even trained with exotic languages, with pre-training initialization, the model still works reasonably well.

Note that in the most challenging case, Exotic Full, where the model does not have any knowledge of both sides, with only 11K parallel pairs for Nl(Dutch)-Pt(Portuguese), the pre-training model still reaches reasonable performance, while the baseline fails to train appropriately. It suggests the pre-train model does learn language-universal knowledge and can transfer to exotic languages easily.

\section{Analysis}
\label{sec: analysis}

\begin{table*}[ht]
\centering
\begin{tabular}{rcccccccccccc}
\toprule

Lang-Pairs& 
\mf{En-Af} &
\mf{En-Ro} &
\mf{En-De} &
\mf{En-Fr} & 

\\

%

Size &
\smf{41K}  & 
\smf{600k}  & 
\smf{4.5M} & 
\smf{40M} & 

\\

 Direction & 
 $\rightarrow$ &$\leftarrow$ &

 $\rightarrow$ &$\leftarrow$ &

 $\rightarrow$ &$\leftarrow$ &

 $\rightarrow$ &$\leftarrow$ \\

\midrule
\baseline &
8.3 & 7.2 & 
30.5 & 29.2 &  
29.3 & 30.8 & 
43.2 & 39.8 

\\

 NA-\method \textit{w/o ft} &
 16.1 &  23.2 &
 24.4 &  33.9 &
22.5 &  30.9 &
 38.6 & 37.3

\\

\midrule

 \method \textit{w/o ft} &
 18.5 &  23.9 &
 25.2 &  34.7 &
 24.2 &  31.2 &
 39.6 & 37.6

\\

 \method &
 31.1 &  27.0 &
  39.0 & 37.4 &
 30.3 &  34.4 &
 44.3 & 45.4  

\\

\bottomrule
\end{tabular}
\caption{
MT performance of \method with and without the RAS technique and fine-tuning strategy. 
\method includes both the RAS technique and fine-tuning strategy. We report tokenized BLEU for this experiment.
``\textit{w/o ft}'' denotes ``without fine-tuning''. 
We also report \method without fine-tuning and RAS to compare with \method without fine-tuning. 
Both RAS and fine-tuning proves effective and essential for \method. 
} 
\label{tab:No-tune-Compare}
\end{table*}

In this section, we conduct a set of analytical experiments to better understand what contributes to performance gains. Three aspects are studied. 
First, we study whether the main contribution comes from pre-training or fine-tuning by comparing the performance of fine-tuning and no-fine-tuning. The results suggest that the performance mainly comes from pre-training, while fine-tuning further boosts the performance. Second, we thoroughly analyze the difference between incorporating RAS at the pre-training phase and pre-training without RAS. The finding shows that incorporating alignment information helps bridge different languages and obtains additional gains.
Lastly, we study the effect of data volume in the fine-tuning phase.

\paragraph{The effects with fine-tuning}.

In the pre-training phase, the model jointly learns from different language pairs. 
To verify whether the gains come from pre-training or fine-tuning, we directly measure the performance without any fine-tuning, which is, in essence, zero-shot translation task. 

We select datasets covering different scales. Specifically, En-Af (41k) from extremely low resource, En-Ro (600k) from low resource, En-De (4.5M) from medium resource, and En-Fr (40M) from rich resource are selected.

As shown in Table \ref{tab:No-tune-Compare} , we find that model without fine-tuning works surprisingly well on all datasets, especially in low resource where we observe model without fine-tuning outperforms randomly initialized baseline model. It suggests that the model already learns well on the pre-training phase, and fine-tuning further obtains additional gains. We suspect that the model mainly tunes the embedding of specific language at the fine-tuning phase while keeping the other model parameters mostly unchanged. Further analytical experiments can be conducted to verify our hypothesis. 

Note that we also report pre-trained model without RAS (NA-\method). For comparison, we do not apply fine-tuning on NA-\method. \method consistently obtains better performance that NA-\method, implying that injecting information at the pre-training phase do improve the performance.

\paragraph{The effectiveness of RAS technique}.

In the pre-training phase, we explicitly incorporate RAS. To verify the effectiveness of RAS, we first compare the performance of \method and \method without RAS.

\begin{table}[ht]
\begin{center}
\scalebox{0.88}{
\begin{tabular}{lcccccc}
\toprule

Lang-Pairs& 

\mf{En-Af} & 
\mf{En-Ro} &
\mf{En-De}

\\

%



 Direction & 
 $\rightarrow$ &$\leftarrow$ &

 $\rightarrow$ &$\leftarrow$ &

 $\rightarrow$ &$\leftarrow$ \\

\midrule


 NA-\method &
 30.6 & 25.4  &
  36.3 &  36.4 &
 27.7 &  33.2

\\




 \method&
 31.1 & 27.0 &
  39.0 &  37.4 &
 30.3 & 34.4

\\

\bottomrule
\end{tabular}
}
\caption{The MT performance of three language pairs with and without alignment information  at pre-training phase. We see consistent performance gains for \method (\method w/ RAS) compared with NA-\method(\method w/o RAS).
 }
\label{tab:No-align-Compare}
\end{center}
\end{table}

As illustrated in Table \ref{tab:No-align-Compare}, We find that utilizing RAS in the pre-training phase consistently helps improve the performance in datasets with different scales, obtaining gains up to 2.5+ BLEU points. 

\begin{figure}[ht] \centering
\subfigure[en-zh w/o RAS]{
  \includegraphics[width=.46\linewidth]{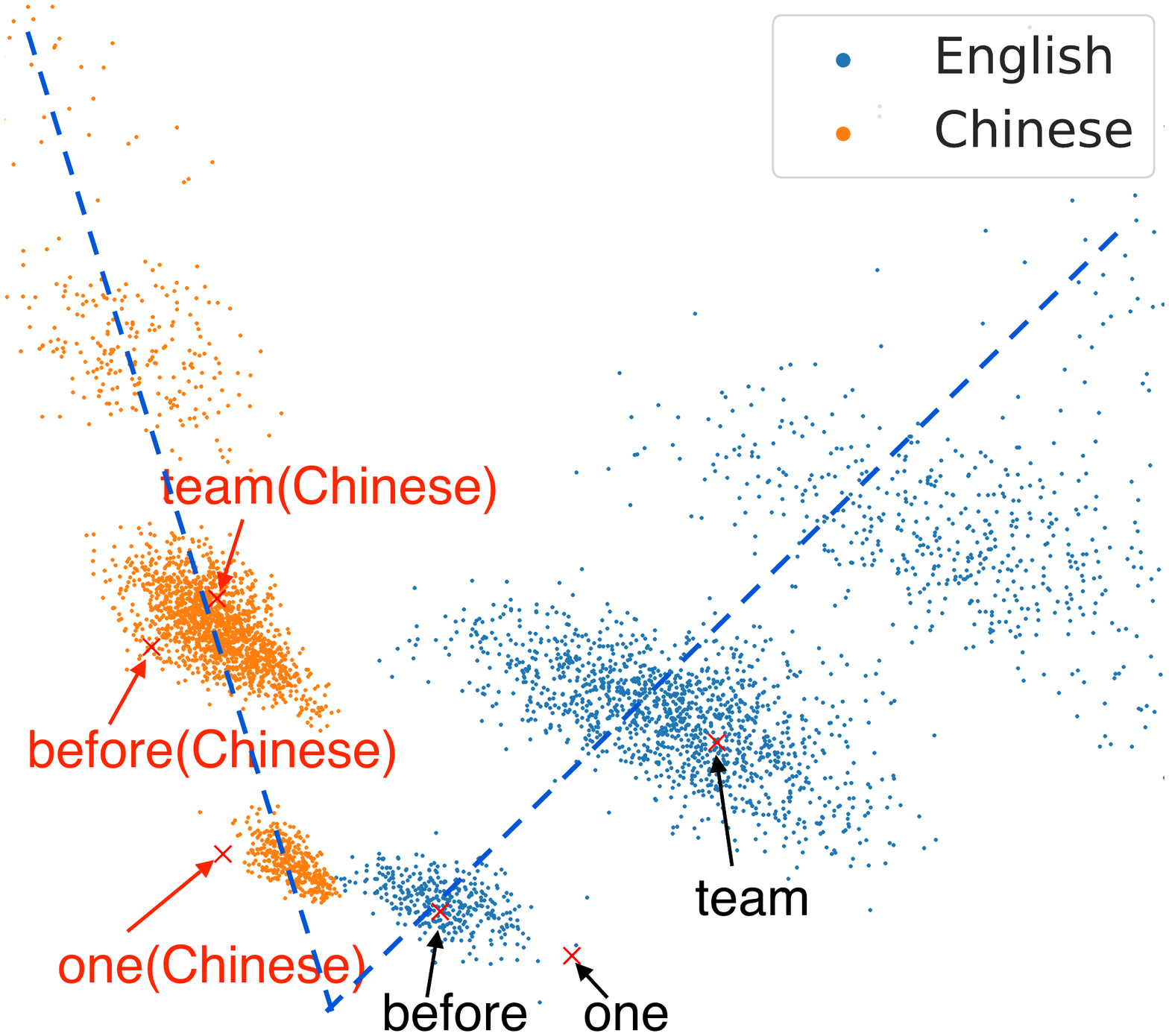}  
  \label{fig:struct1}
  }
  \subfigure[en-zh w/ RAS]{
  \includegraphics[width=.46\linewidth]{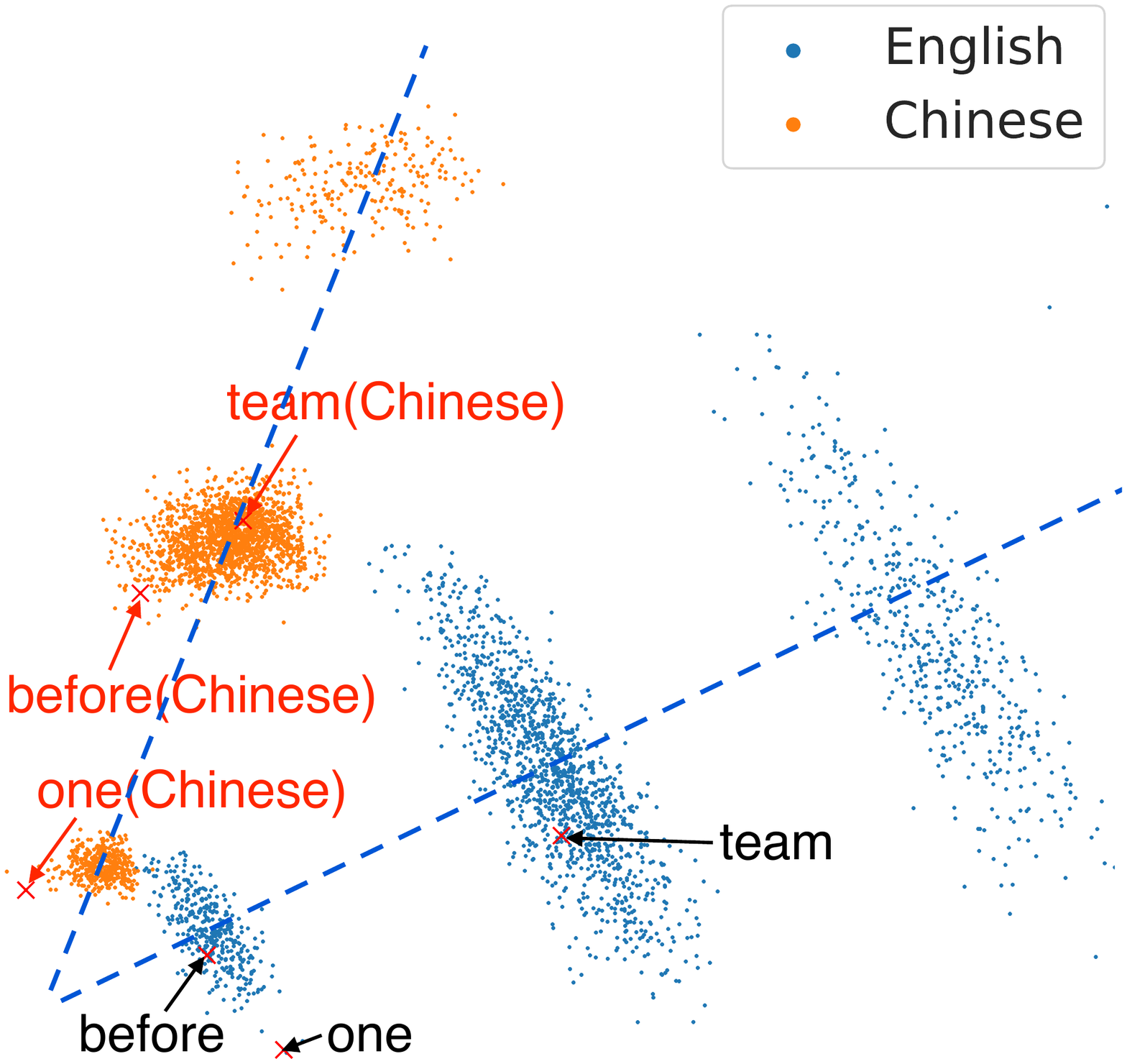}  
   \label{fig:struct2}
  }
  \subfigure[en-af w/o RAS]{
  \includegraphics[width=.46\linewidth]{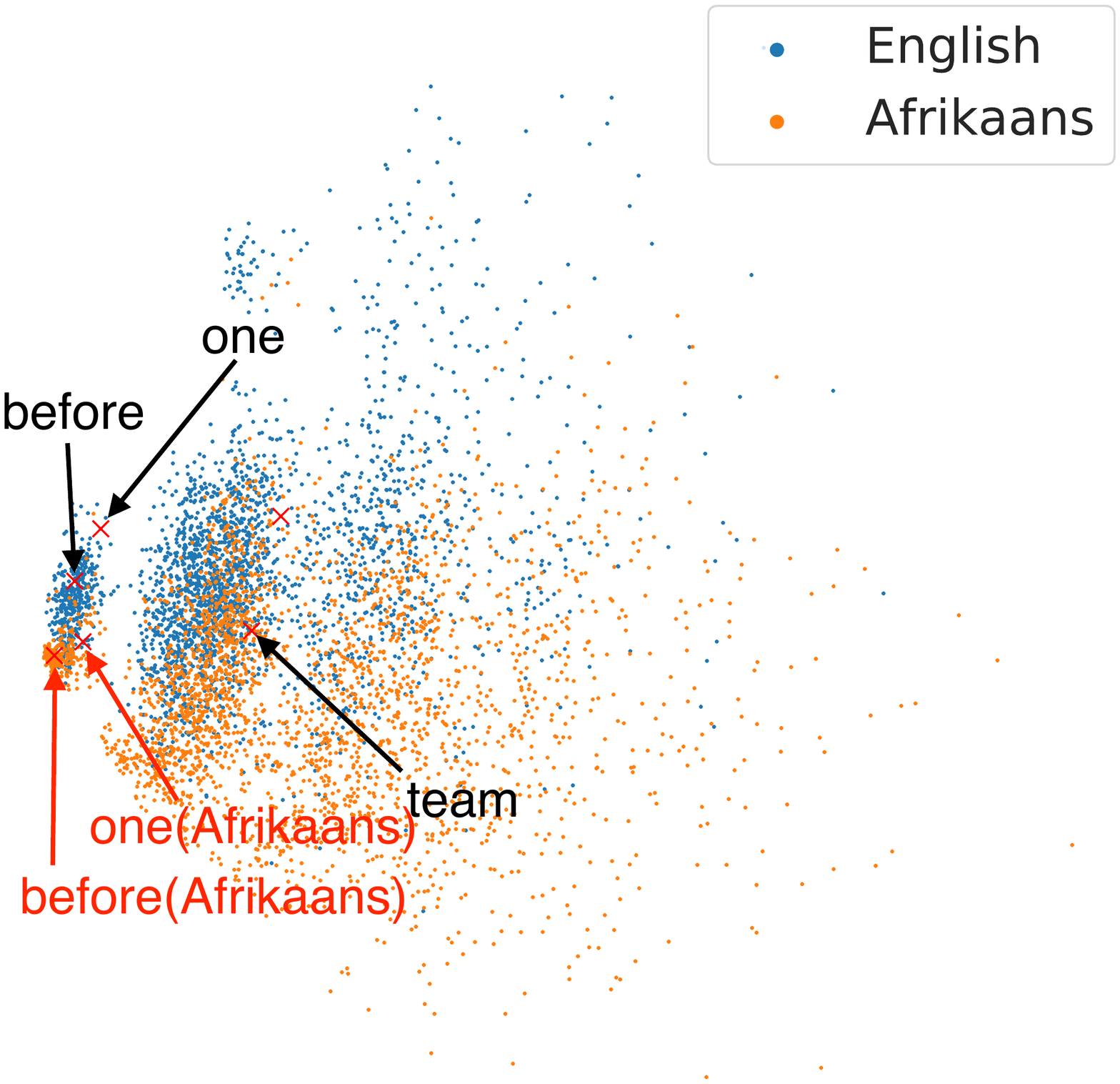}  
   \label{fig:struct3}
  }
  \subfigure[en-af w/ RAS]{
  \includegraphics[width=.46\linewidth]{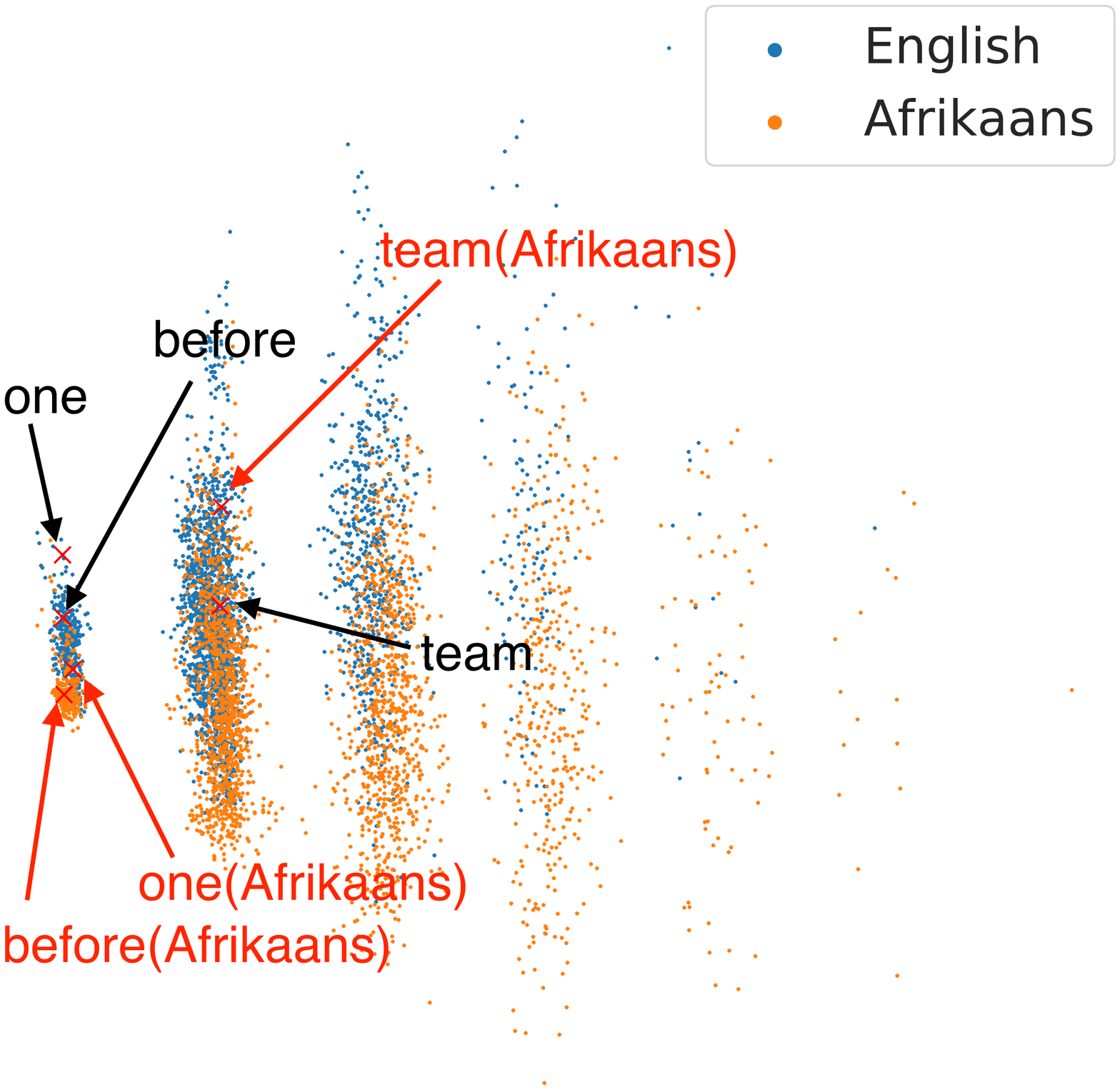}  
   \label{fig:struct4}
  }
\caption{Visualization of Word Embedding from NA-\method (w/o RAS) vs \method (w/ RAS). For both similar language pairs and dissimilar language pairs that have no lexical overlap, the word embedding distribution becomes closer after RAS.}
\label{fig:struct}
\end{figure}

\begin{figure}[ht]
    \centering
    \includegraphics[width=1.03\linewidth]{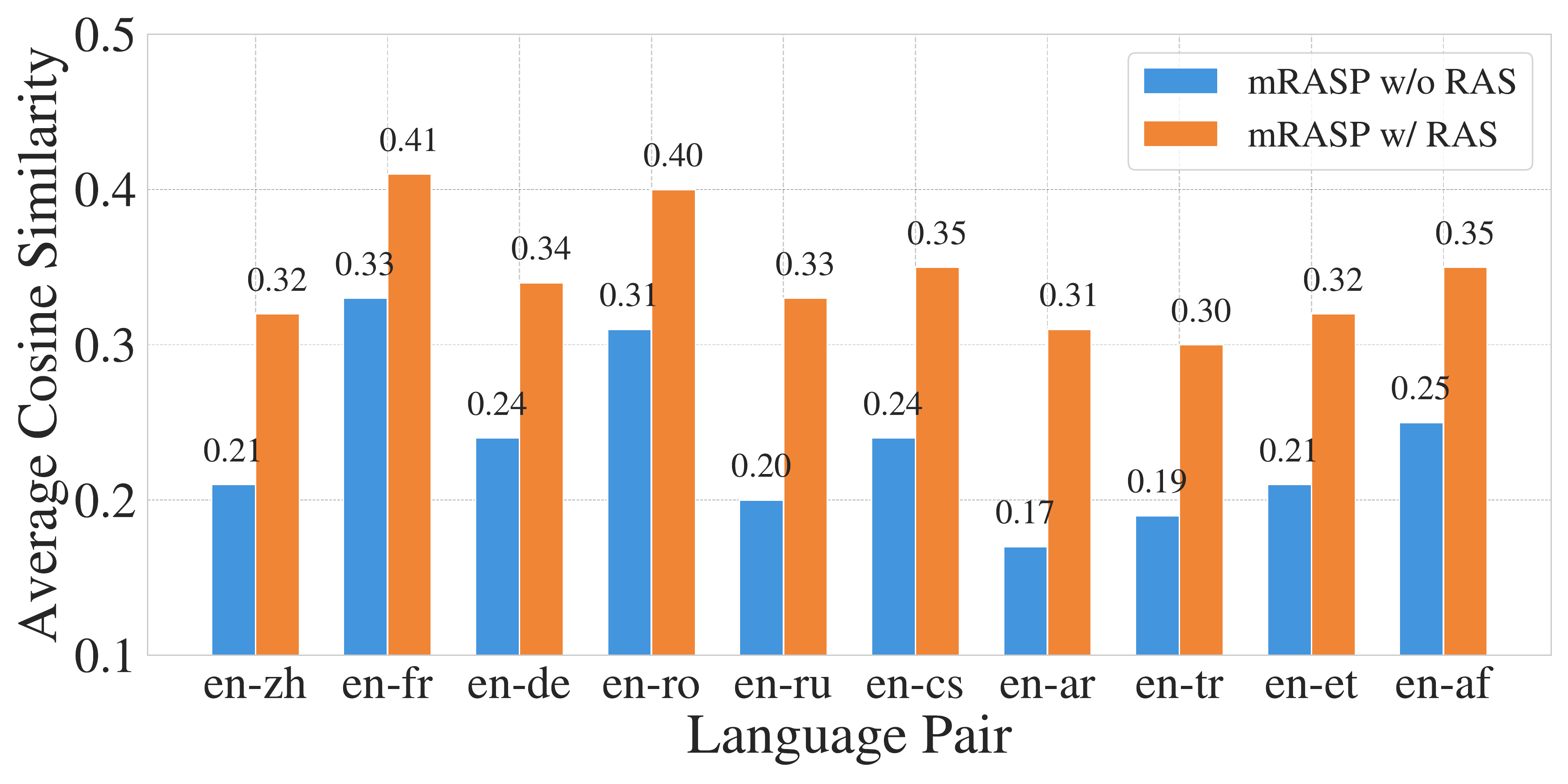}
    \caption{Average cosine similarity NA-\method (\method w/o RAS) vs \method (\method w/ RAS). The similarity increases after applying the RAS technique, which explains the effectiveness of RAS. }
    \label{fig:cos-sim}
\end{figure}

To verify whether the semantic space of different languages draws closer after adding alignment information quantitatively, we calculate the average cosine similarity of words with the same meaning in different languages. We choose the top frequent 1000 words according to MUSE dictionary. Since words are split into subwords through BPE, we simply add all subwords constituting the word.   As illustrated in Figure \ref{fig:cos-sim}, we find that for all pairs in the Figure, the average cosine similarity increases by a large margin after adding RAS, suggesting the efficacy of alignment information in bridging different languages. It is worth mentioning that the increase does not only happen on similar pairs like En-De, but also on dissimilar pairs like En-Zh.

To further illustrate the effect of RAS on semantic space more clearly, we use PCA (Principal Component Analysis)  to visualize the word embedding space. We plot En-Zh as the representative for dissimilar pairs and En-Af for similar pairs. More figures can be found in the Appendix.

As illustrated in Figure \ref{fig:struct}, we find that for both similar pair and dissimilar pair, the overall word embedding distribution becomes closer after RAS. For En-Zh, as the dashed lines illustrate, the angle of the two word embedding spaces becomes smaller after RAS. And for En-Af, we observe that the overlap between two space becomes larger. We also randomly plot the position of three pairs of words, with each pair has the same meaning in different languages. 

\paragraph{Fine-tuning Volume}
\begin{figure}[ht]
    \centering
    \includegraphics[width=.48\textwidth]{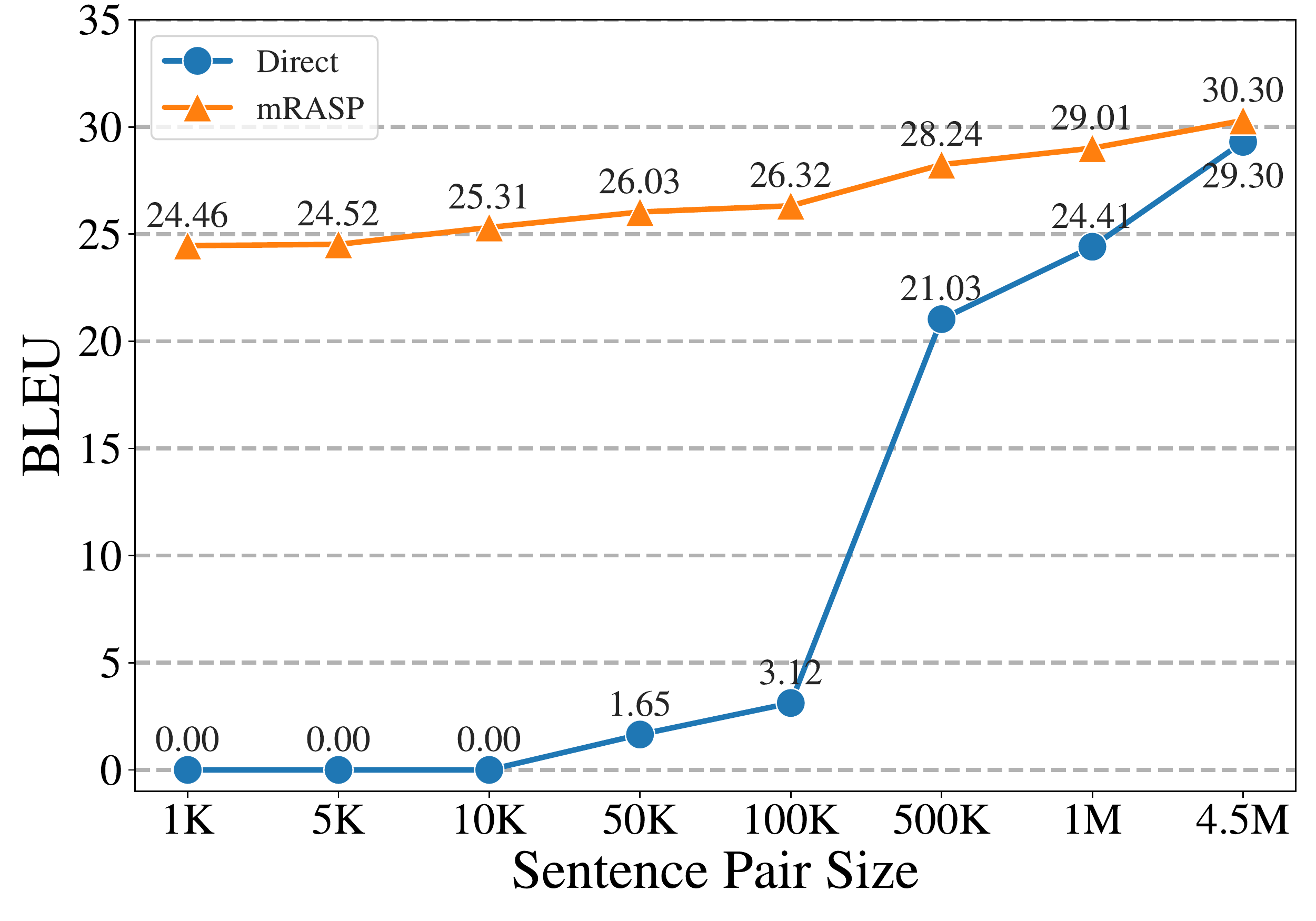}
    \caption{Performance curves for En$\rightarrow$De along with the size of parallel pairs. With \method pre-trained model, the fine-tuned down-stream MT model is able to obtain descent translation performance even when there is very small corpus to train. }
    \label{fig:compare}
\end{figure}

To study the effect of data volume in the fine-tuning phase, we randomly sample 1K, 5K, 10K, 50K, 100K, 500K, 1M datasets from the full En-De corpus (4.5M). We fine-tune the model with the sampled datasets, respectively. Figure \ref{fig:compare} illustrates the trend of BLEU with the increase of data volume. With only 1K parallel pairs, the pre-trained model works surprisingly well, reaching 24.46. As a comparison, the model with random initialization fails on this extremely low resource. 
With only 1M pairs, \method reaches comparable results with baseline trained on 4.5M pairs.

With the size of dataset increases, the performance of the pre-training model consistently increases. 
While the baseline does not see any improvement until the volume of the dataset reaches 50K. The results confirm the remarkable boosting of \method on low resource dataset.

\section{Related Works}
\label{sec:related}

\paragraph{ Multilingual NMT}  aims at taking advantage of multilingual data to improve NMT for all languages involved, which has been extensively studied in a number of papers such as ~\citet{dong2015multi,DBLP:journals/tacl/JohnsonSLKWCTVW17,lu2018neural,rahimi2019massively,tan2019multilingual}. 
The most related work to \method is ~\citet{rahimi2019massively}, which performs extensive experiments in
training massively multilingual NMT models.  
They show that multilingual many-to-many models are effective in low resource settings.
Inspired by their work, we believe that the translation quality of  low-resource language pairs may improve when trained together with rich-resource ones.
However, we are different in at least two aspects: \begin{inparaenum}[\it a)]
    \item Our goal is to find the best practice of a single language pair with multilingual pre-training. Multilingual NMT usually achieves inferior accuracy compared with its counterpart, which trains an individual model for each language pair when there are dozens of language pairs.
    \item Different from multilingual NMT, \method can obtain improvements with rich-resource language pairs, such as English-Frence.
\end{inparaenum}

\paragraph{Unsupervised Pretraining}  has
significantly improved the state of the art in natural language understanding from word embedding~\cite{mikolov2013distributed,pennington2014glove}, pretrained contextualized representations~\cite{DBLP:conf/naacl/PetersNIGCLZ18,radford2019language,DBLP:conf/naacl/DevlinCLT19} and sequence to sequence pretraining~\cite{DBLP:conf/icml/SongTQLL19}. 
It is widely accepted that one of the most important factors for the success of unsupervised pre-training is the scale of the data. 
The most successful efforts, such as RoBERTa, GPT, and BERT, highlight the importance of scaling the amount of data. 
Following their spirit,  we show that with massively multilingual pre-training, more than 110 million sentence pairs, \method can significantly boost the performance of the downstream NMT tasks. 

On parallel, there is a bulk of work on unsupervised cross-lingual representation.  
Most traditional studies show that cross-lingual
representations can be used to improve the quality of monolingual representations. 
\citet{mikolov2013exploiting} first introduces dictionaries to align
word representations from different languages. 
A series of follow-up studies focus on aligning the word representation across languages~\cite{xing2015normalized,ammar2016massively,smith2017offline,conneau2017word}. 
Inspired by the success of BERT, ~\citet{DBLP:conf/nips/ConneauL19} introduced XLM - masked language models trained on multiple languages, as a way to leverage parallel data and obtain impressive empirical results on the cross-lingual natural language inference
(XNLI) benchmark and unsupervised NMT\cite{DBLP:conf/acl/SennrichHB16,DBLP:conf/iclr/LampleCDR18,DBLP:journals/corr/abs-2002-02955}. 
~\citet{huang2019unicoder} extended XLM with multi-task learning and proposed a universal language encoder. 

Different from these works,  
\begin{inparaenum}[\it a)]
    \item \method is actually a multilingual sequence to sequence model which is more desirable for  NMT pre-training;
    \item \method  introduces alignment regularization to bridge the sentence representations across languages.  
\end{inparaenum}

\section{Conclusion}
\label{sec:conclusion}
In this paper, we propose a multilingual neural machine translation pre-training model (\method). To bridge the semantic space between different languages, we incorporate word alignment into the pre-training model. Extensive experiments are conducted on different scenarios, including low/medium/rich resource and exotic corpus, demonstrating the efficacy of \method. We also conduct a set of analytical experiments to quantify the model, showing that the alignment information does bridge the gap between languages as well as boost the performance. We leave different alignment approaches to be explored in the future. In future work, we will pre-train on larger corpus to further boost the performance.

\section*{Acknowledgments}
We would like to thank the anonymous reviewers for their valuable comments.
We would also like to thank Liwei Wu, Huadong Chen, Qianqian Dong, Zewei Sun, and Weiying Ma for their useful suggestion and help with experiments.

\bibliographystyle{acl_natbib}
\bibliography{paper}

\newpage

\appendix
\section{Appendices}
\label{sec:appendix}

\subsection{Visualization of Word Embedding}

\begin{figure*}[!h]
\subfigure[en-ro w/o RAS]{
  \includegraphics[width=.5\linewidth]{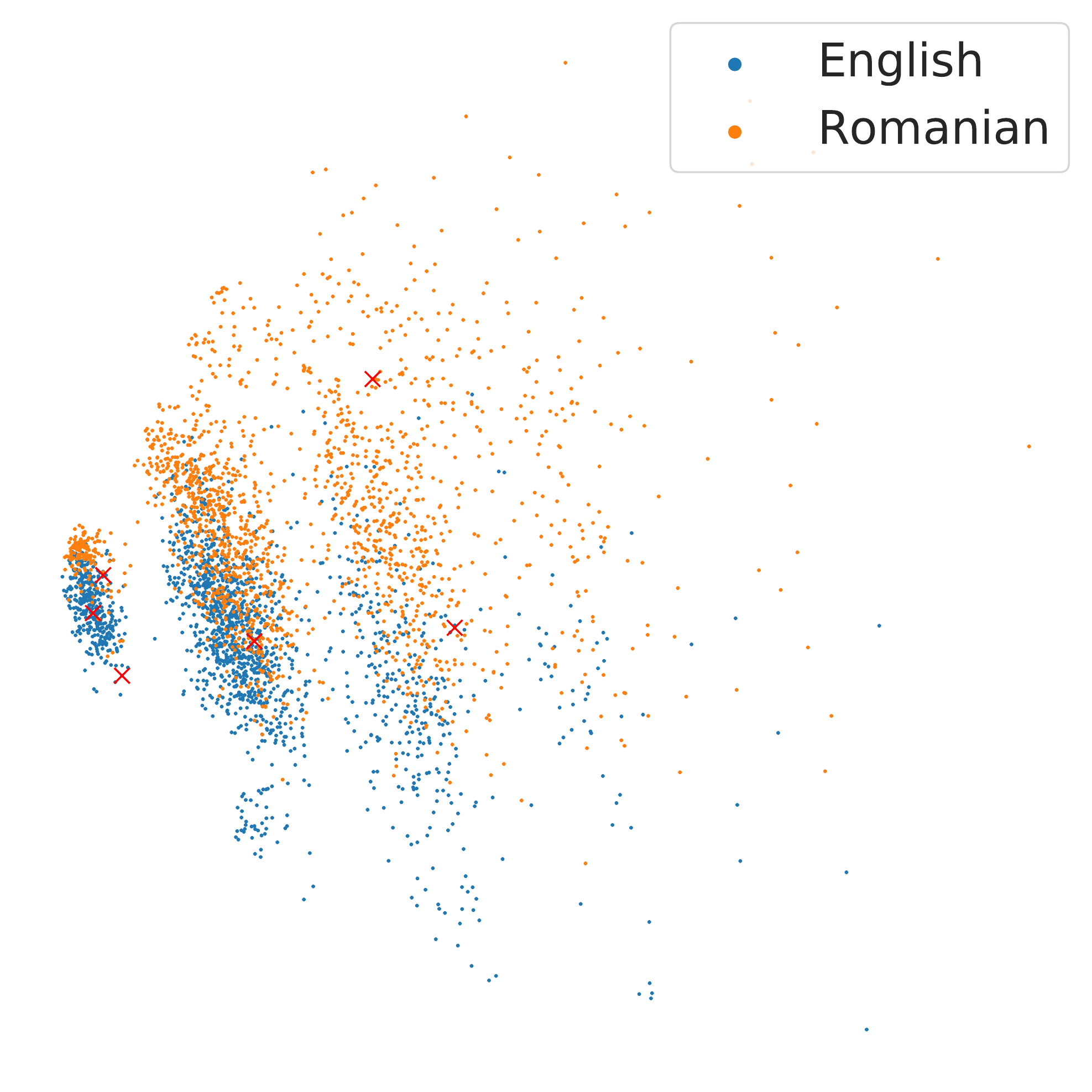}  
  \label{fig:struct1-ro}
  }
  \subfigure[en-ro w/ RAS]{
  \includegraphics[width=.5\linewidth]{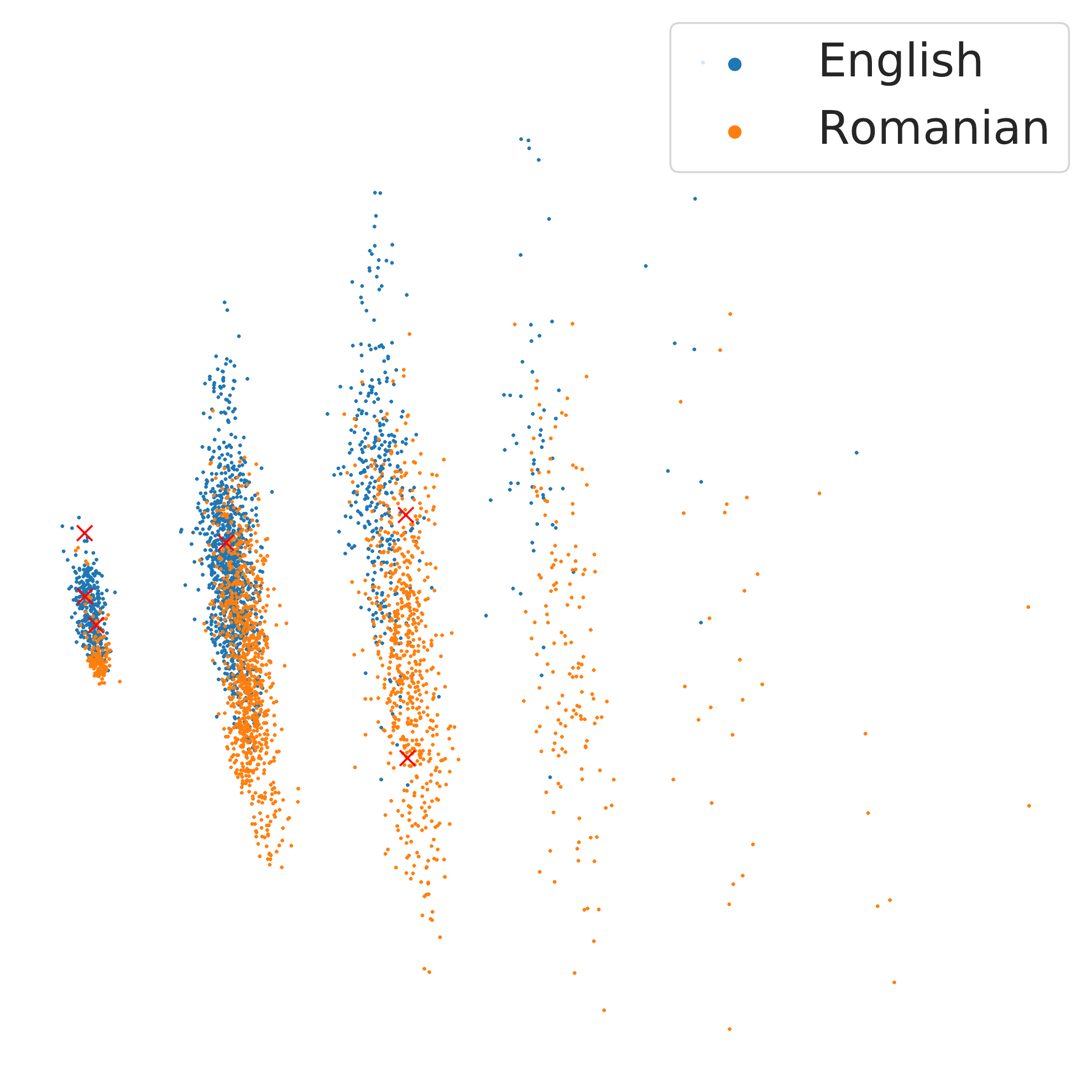}  
   \label{fig:struct2-ro}
  }
\caption{Visualization of Word Embedding from \method w/o RAS vs \method w/ RAS for English-Romanian}
\label{fig:struct-ro}
\end{figure*}

\begin{figure*}[!h]
\subfigure[en-ar w/o RAS]{
  \includegraphics[width=.5\linewidth]{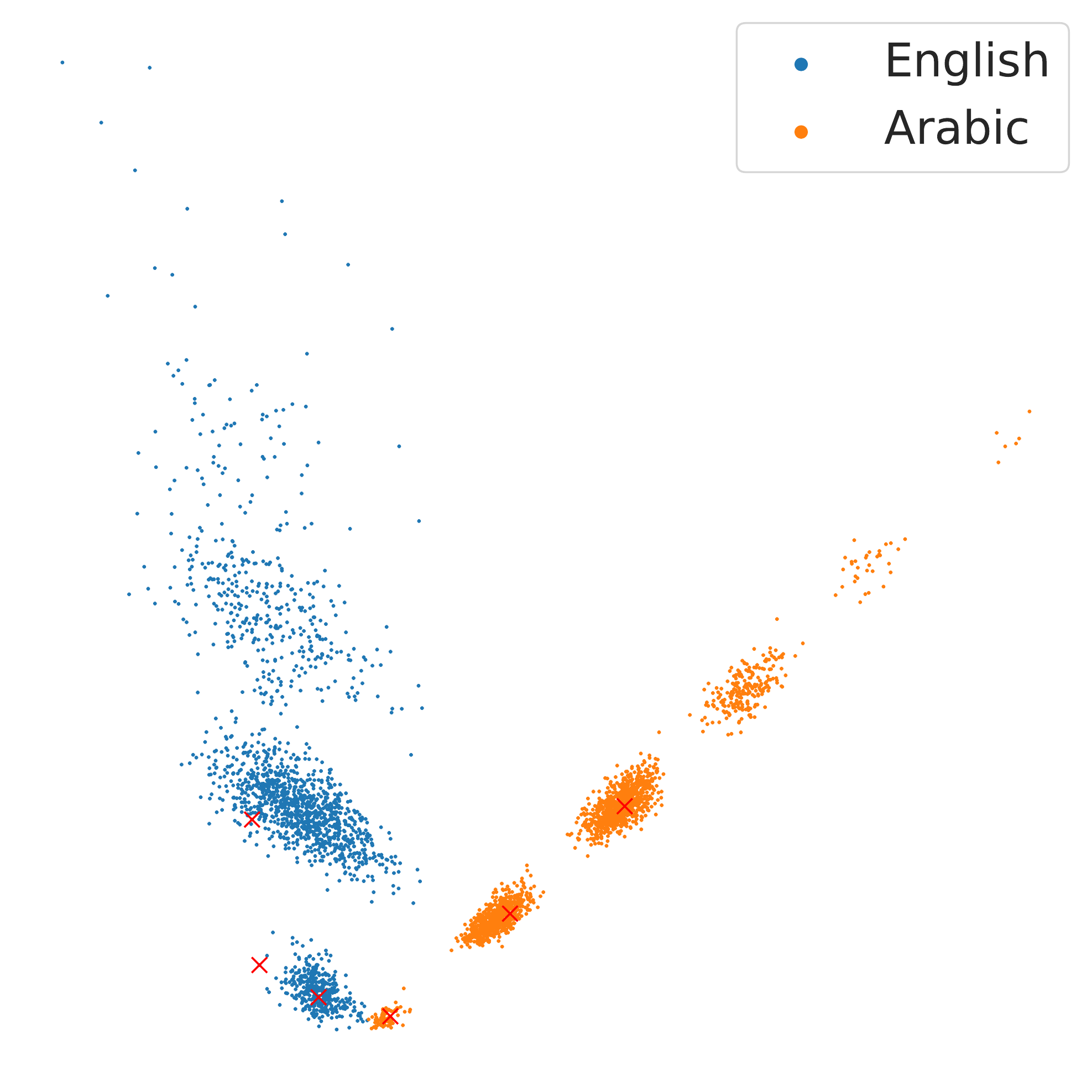}  
  \label{fig:struct1-ar}
  }
  \subfigure[en-ar w/ RAS]{
  \includegraphics[width=.5\linewidth]{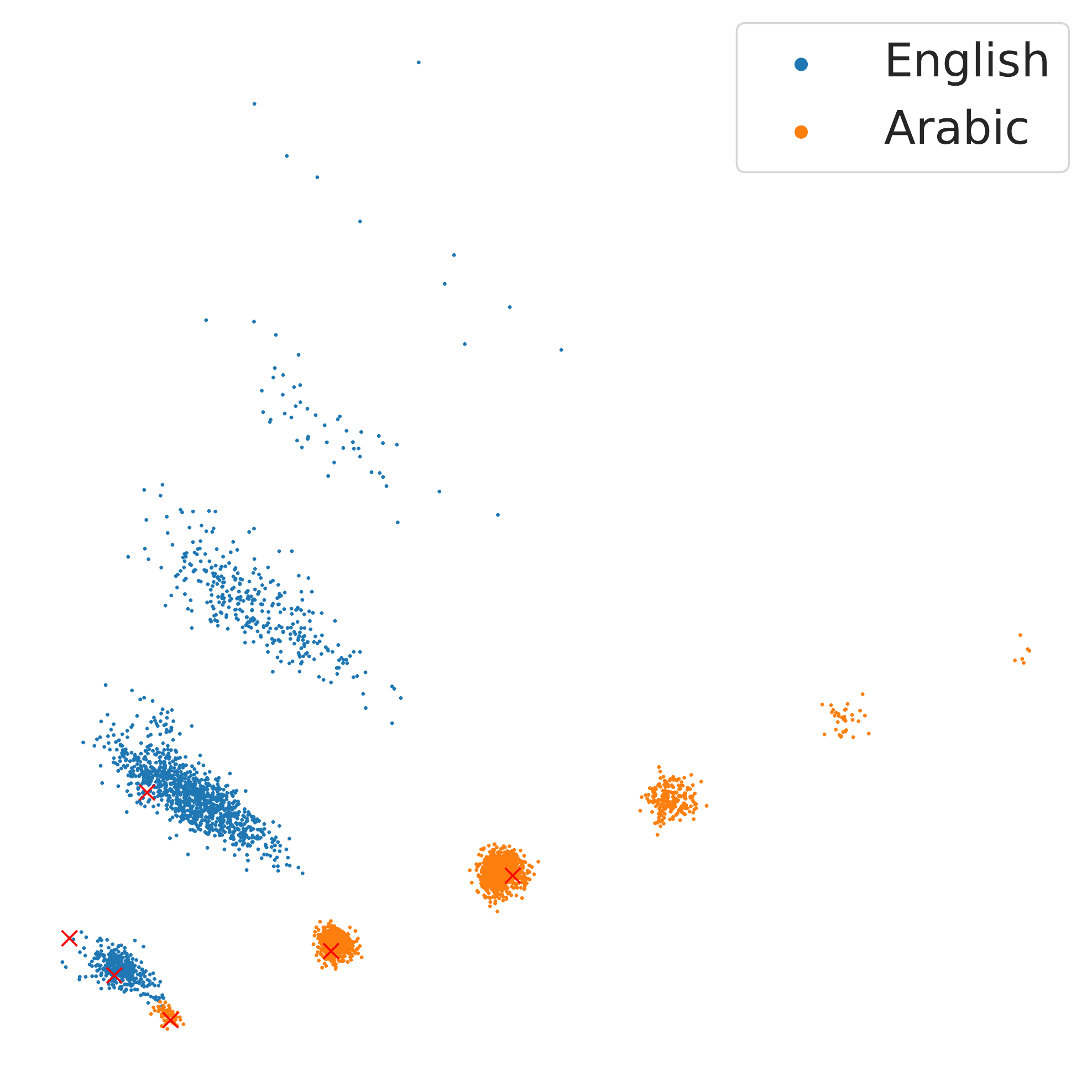}  
   \label{fig:struct2-ar}
  }
\caption{Visualization of Word Embedding from \method w/o RAS vs \method w/ RAS for English-Arabic}
\label{fig:struct-ar}
\end{figure*}

\begin{figure*}[!h]
\subfigure[en-tr w/o RAS]{
  \includegraphics[width=.5\linewidth]{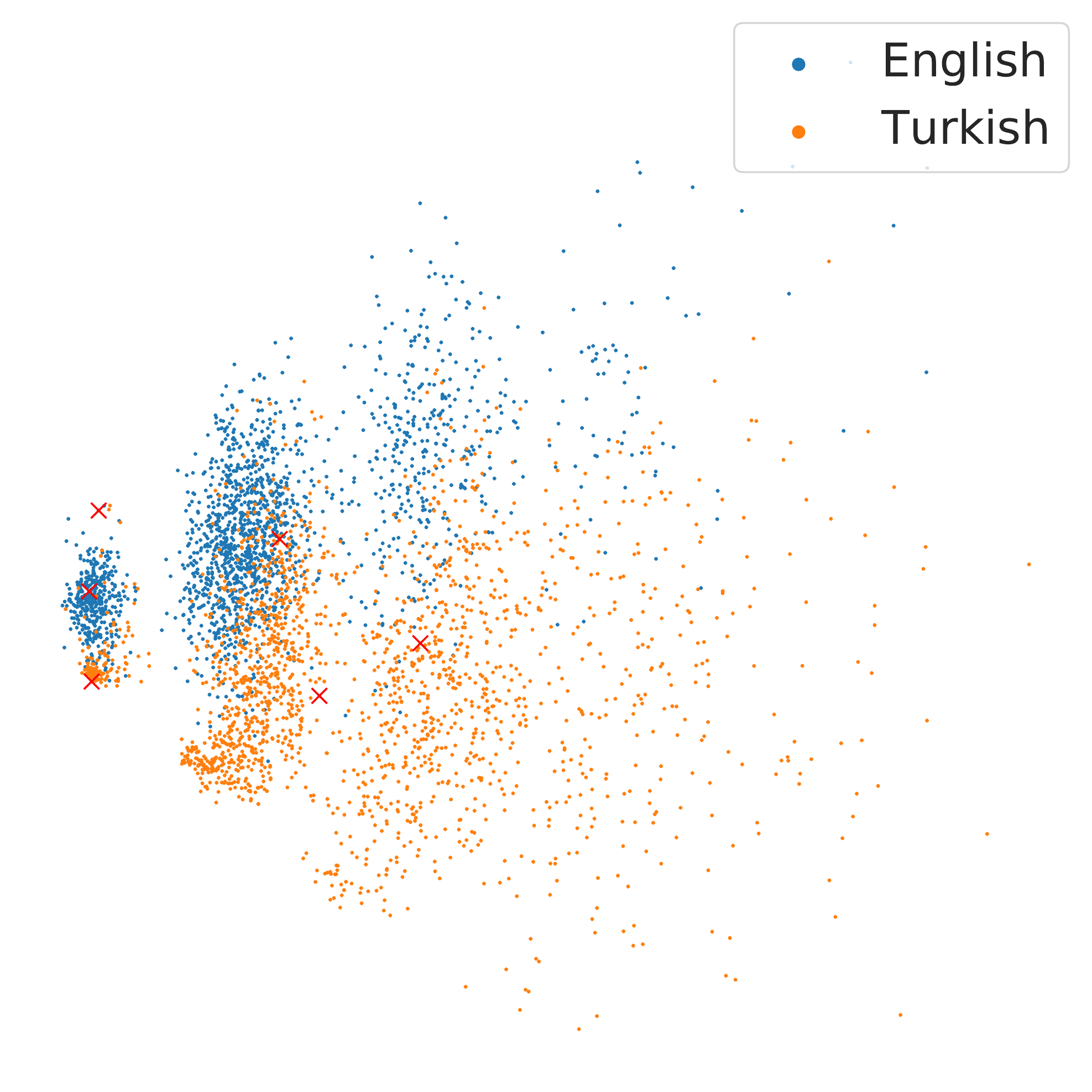}  
  \label{fig:struct1-tr}
  }
  \subfigure[en-tr w/ RAS]{
  \includegraphics[width=.5\linewidth]{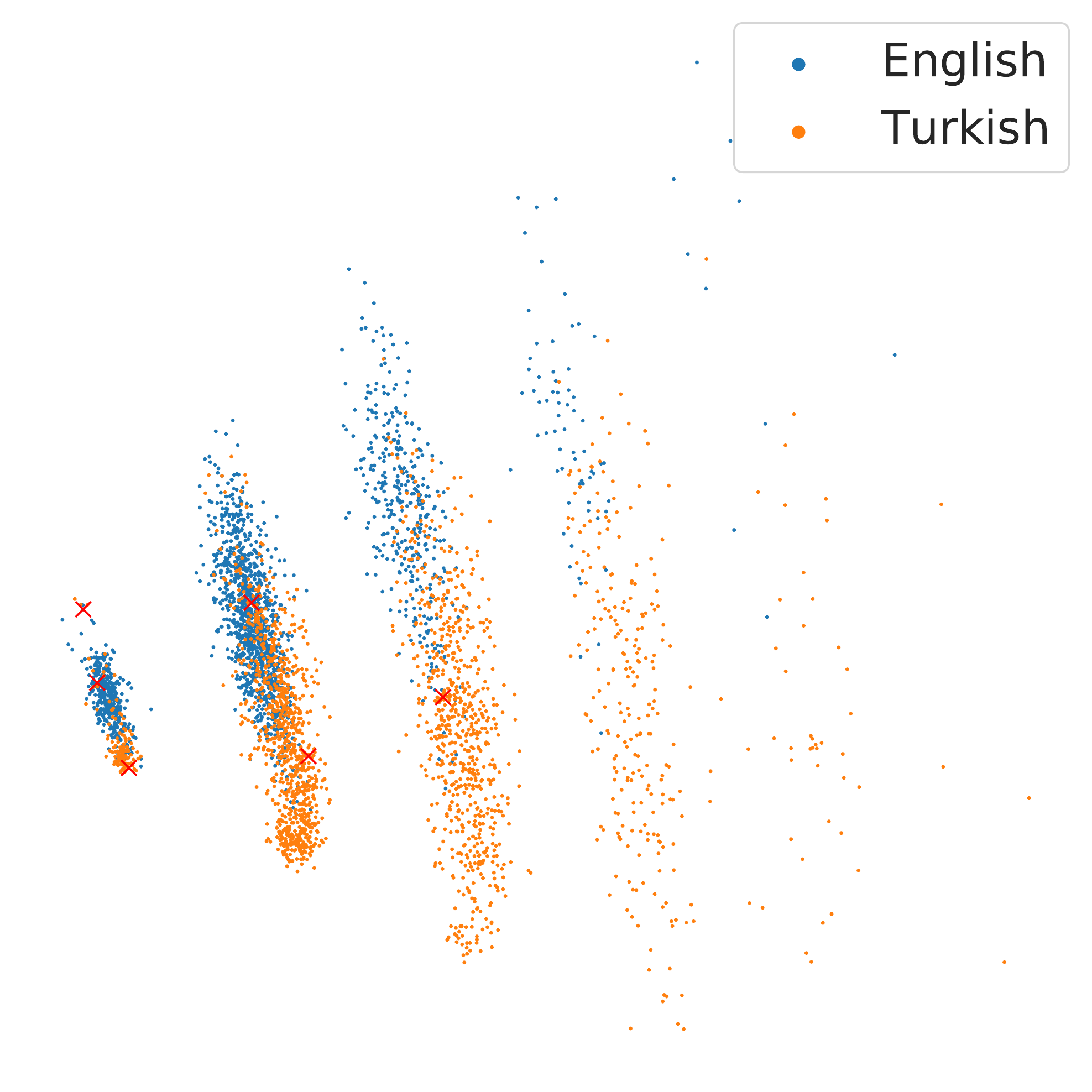}  
   \label{fig:struct2-tr}
  }
\caption{Visualization of Word Embedding from \method w/o RAS vs \method w/ RAS for English-Turkish}
\label{fig:struct-tr}
\end{figure*}

\begin{figure*}[!h]
\subfigure[en-de w/o RAS]{
  \includegraphics[width=.5\linewidth]{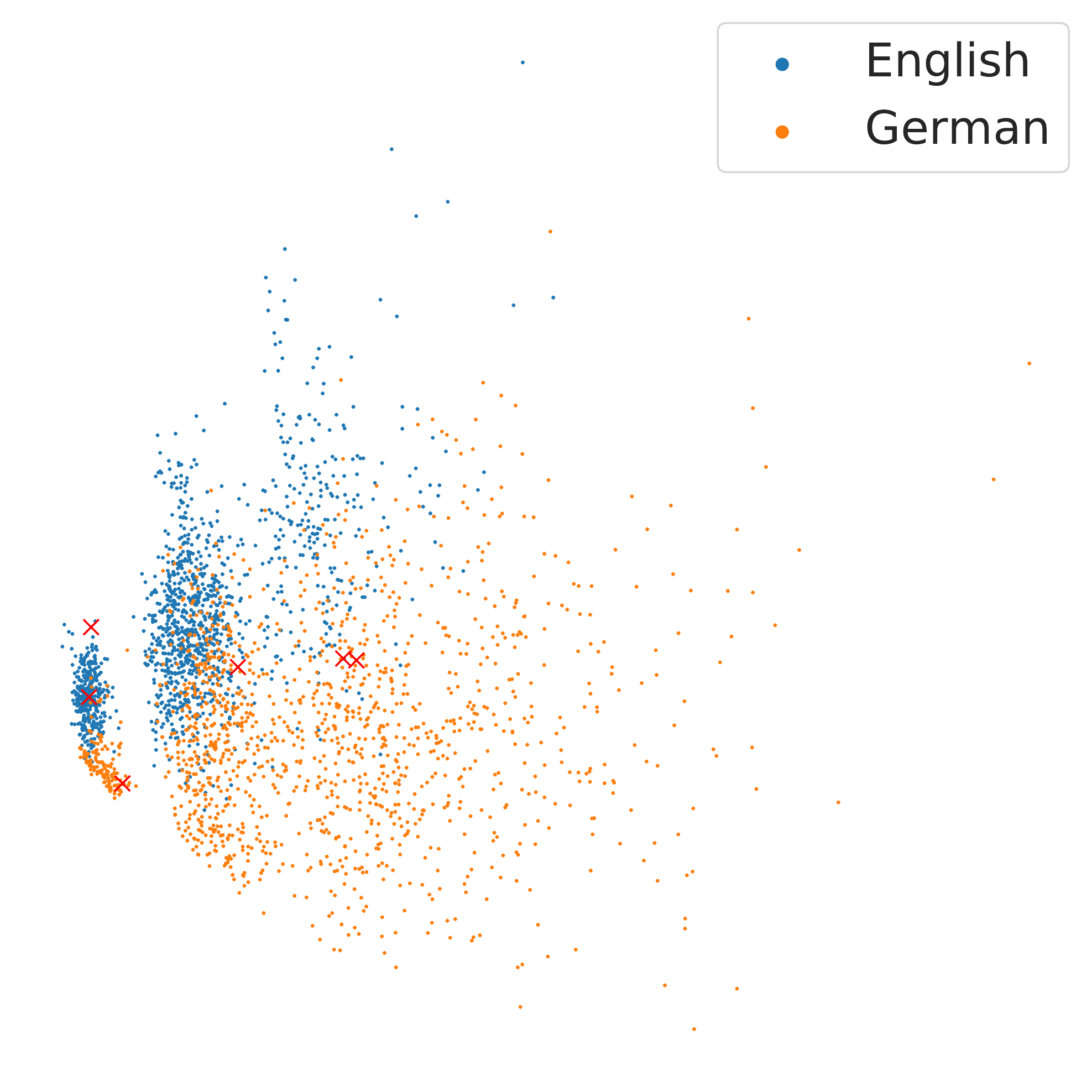}  
  \label{fig:struct1-de}
  }
  \subfigure[en-de w/ RAS]{
  \includegraphics[width=.5\linewidth]{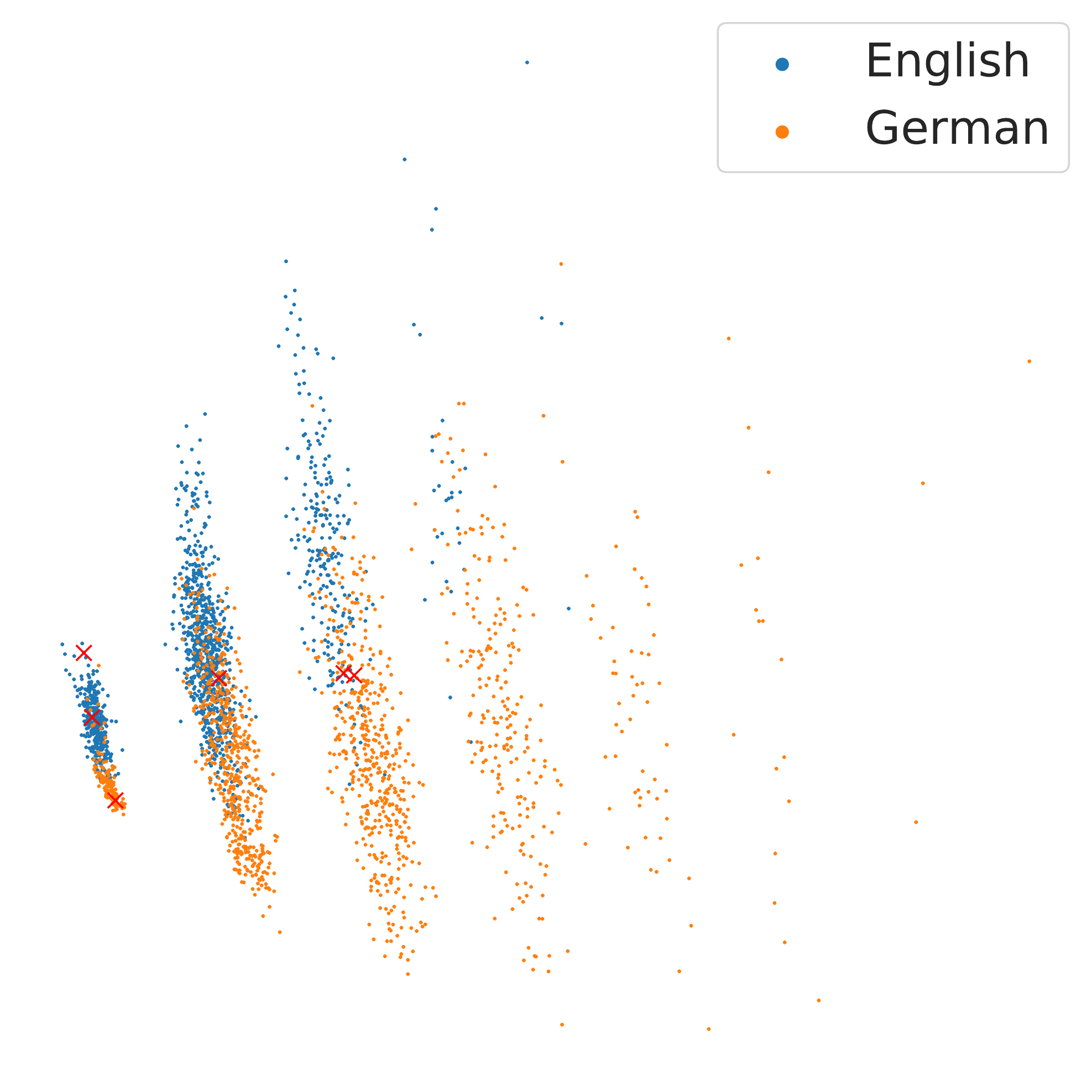}  
   \label{fig:struct2-de}
  }
\caption{Visualization of Word Embedding from \method w/o RAS vs \method w/ RAS for English-German}
\label{fig:struct-de}
\end{figure*}

In addition to visualization of En-Zh and En-Af presented in main body of paper, we also plot visualization of En-Ro, En-Ar, En-Tr and En-De. As shown in Figure \ref{fig:struct-ro},\ref{fig:struct-ar},\ref{fig:struct-tr},\ref{fig:struct-de}, the overall word embedding distribution becomes closer after RAS.

\subsection{Case Study}

\begin{CJK}{UTF8}{gkai}
\begin{table*}[t]
\begin{center}
\begin{tabular}{ll}
\toprule
En$\rightarrow$ Fr &  \\
\midrule
Src & An investigation is under way to find the cause of the fire. \\
Ref & Une enquête est en cours pour trouver la cause de cet incendie. \\
Direct & enquête est en cours pour déterminer la cause de l' incendie. \\
mRASP & Une enquête est en cours pour trouver la cause de l' incendie. \\

\midrule

En$\rightarrow$ Zh &  \\
\midrule

Src & and for the middle class. \\

Ref & 对中产阶级而言。 \\
Direct & 还有中产阶级。 \\
mRASP & 对中产阶级而言。 \\

\midrule

Fr$\rightarrow$ Zh &  \\
\midrule

Src & Ordre du jour provisoire de la 7424e séance ( privée ) du Conseil \\
Ref & 安全理事会第7424次(闭门)会议临时议程 \\
Direct & 事实上，国际货币基金组织的国际货币基金组织（IMF） \\
mRASP & 安理会第7424次（非公开）会议临时议程 \\

\midrule

Nl$\rightarrow$ Pt &  \\
\midrule
Src & de notulen van de vergadering van donderdag 21 september zijn rondgedeeld. \\
Ref & a acta da sessão de quinta feira, 21 de setembro de 2000 , já foi distribuída. \\
Direct & Os governos, os líderes mundiais dos seus próprios. \\
mRASP & As notícias da reunião do dia 21 de Setembro foram partilhadas. \\

\bottomrule
\end{tabular}
\caption{Case Study}
\label{case-en2fr}
\end{center}
\end{table*}

\end{CJK}

\subsection{Results on public testsets}

\begin{table*}[htb]
\begin{center}
\begin{tabular}{rcccccccccccc}
\toprule
&\multicolumn{7}{c}{\textbf{Extremely Low Resource ($<$100k)}} \\
\midrule

Lang-Pairs& 
\mf{En-Be (opus-100)} &
\mf{En-My (opus-100)} &
\mf{En-Af (opus-100)} &
\mf{En-Eo (opus-100)} & 
Avg

\\

%

Size &
\smf{20K}  & 
\smf{29k}  & 
\smf{41K} & 
\smf{67K} & 

\\

Direction & 
 $\rightarrow$ &$\leftarrow$ &

 $\rightarrow$ &$\leftarrow$ &

 $\rightarrow$ &$\leftarrow$ &

 $\rightarrow$ &$\leftarrow$ \\

\midrule
\baseline &
1.5 & 0.6 & 
0 & 0.2 &  
6.1 & 5.8 & 
11.7 & 10.1 & 
4.5
\\

 \method &
\bf 13.4 & \bf 16.2 &
\bf 1.8 & \bf 7.3 &
\bf 21.3 & \bf 25.5 &
\bf 30.7 & \bf 32.9 &
18.6

\\

 $\Delta$ &
 +11.9 & +15.6 &
 +1.8 & +7.1  &
 +15.2 & +19.7 &
 +19.0 & +22.8 &
 \textbf{+14.1}

\\

\midrule
\midrule

&\multicolumn{7}{c}{\textbf{Low Resource (100k$\sim$1m)}} \\
\midrule

Lang-Pairs &
\mf{En-He} &
\mf{En-Tr (wmt2016)} &
\mf{En-Ro} &
\mf{En-Cs (wmt2016)} &
Avg
\\

Size &
\smf{335K} & 
\smf{388K} & 
\smf{600K} & 
\smf{978K}
\\

 Direction & 
 $\rightarrow$ &$\leftarrow$ &

 $\rightarrow$ &$\leftarrow$ &

 $\rightarrow$ &$\leftarrow$ &

 $\rightarrow$ &$\leftarrow$ \\

\midrule
 \baseline &
/ & / & 
14.1 & 19.2 & 
/ & / & 
21.8 & 26.5 & 
20.4
\\

 \method &
/ & / &  
\bf 17.5 & \bf 22.9 & 
/ & / & 
\bf 24.0 & \bf 30.9 & 
23.8

\\

$\Delta$ &
/ & / &
+3.4 & +3.7  &
/ & / &
+2.2 & +4.4 &
\textbf{+3.4}

\\

\midrule
\midrule
&\multicolumn{7}{c}{\textbf{Medium Resource (1m$\sim$10m)}} \\
\midrule

Lang-Pairs & 
\mf{En-Ar} &
\mf{En-Et (wmt2018)} &
\mf{En-Bg (opus-100)} &
\mf{En-De} &
Avg
\\

Size &
\smf{1.2M}  & 
\smf{2.3M} & 
\smf{3.1M} & 
\smf{4.5M}  

\\

Direction & 
 $\rightarrow$ &$\leftarrow$ &

 $\rightarrow$ &$\leftarrow$ &

 $\rightarrow$ &$\leftarrow$ &

 $\rightarrow$ &$\leftarrow$ \\

\midrule
 \baseline &
/ & / & 
20.2 & 24.5 & 
24.2 & 26.2 & 
/ & / &  
23.8
\\

 \method &
/ & / & 
\bf 21.9 & \bf 28.2 & 
\bf 25.2 & \bf 27.5 & 
 / & / & 
25.7
\\

$\Delta$ &
/ & / &
+1.7 & +3.7 &
+1.0 & +1.3 &
/ & / &
\textbf{+1.9}

\\

\bottomrule
\end{tabular}
\caption{Fine-tuning performance on \emph{extremely low} / \emph{low} / \emph{medium} resource machine translation settings on public testsets.}
\label{tab:LowRpub}
\end{center}
\end{table*}

\subsection{Data Description}
As listed in Table \ref{table-data-description}, we collect 32 English-centric language pairs, resulting in a total pairs of 110M. The parallel corpus are from various source, ted, wmt,
europarl, paracrawl, opensubtitles and qed.

\begin{table*}[htb]
\rowcolors{2}{lightgray}{}
\begin{tabular}{lrrrrrrrrr}
\toprule
 {\textbf{Lang}}&
  {\textbf{Ted}}    & {\textbf{Euro}} & {\textbf{Qed}}    & {\textbf{Ops}} & {\textbf{WMT}}   & {\textbf{Para}} & {\textbf{Others}} & {\textbf{Sum}}  \\

\midrule
Af   & -      & -        & -      & 42429         & -        & -         &        -            & 42429     \\ 
Ar   & 214111 & -        & -      & 1000788       & -        & -         &          -                  & 1214899   \\ 
Be   & 4509   & -        & 21080  & -             & -        & -         &          -                  & 25589     \\ 
Bg   & 174444 & 406934   & -      & -             & -        & 2586277   &          -                  & 3167655   \\ 
Cs   & 103093 & -        & -      & -             & 838037   & -         &            -                & 941130    \\ 
De   & 167888 & -        & -      & -             & 4590101  & -         &           -                 & 4757989   \\ 
El   & 134327 & 1235976  & -      & -             & -        & -         &            -                & 1370303   \\ 
Eo   & 6535   & -        & -      & 61043         & -        & -         &           -                 & 67578     \\ 
Es   & 196026 & 1965734  & -      & -             & -        & -         &          -                  & 2161760   \\ 
Et   & 10738  & -        & -      & -             & 2176827  & 132522    &          -                  & 2320087   \\ 
Fi   & 24222  & 1924942  & -      & -             & 2078670  & -         &          -                  & 4027834   \\ 
Fr   & 192304 & -        & -      & -             & 39816621 & -         & 19870 & 40028795  \\ 
Gu   & -      & -        & -      & -             & 11671    & -         &          -                  & 11671     \\ 
He   & 211819 & -        & -      & 123692        & -        & -         &         -                   & 335511    \\ 
Hi   & 18798  & -        & -      & -             & -        & -         & 1555738                    & 1574536   \\ 
It   & 204503 & 1909115  & -      & -             & -        & -         &       -                     & 2113618   \\ 
Ja   & 204090 & -        & -      & 1872100       & -        & -         &         -                   & 2076190   \\ 
Ka   & 13193  & -        & -      & 187411        & -        & -         &        -                    & 200604    \\ 
Kk   & 3317   & -        & -      & -             & 124770   & -         &         -                   & 128087    \\ 
Ko   & 205640 & -        & -      & 1270001       & -        & -         &         -                   & 1475641   \\ 
Lt   & 41919  & -        & -      & -             & 2342917  & -         &        -                    & 2384836   \\ 
Lv   & -      & -        & -      & -             & 4511715  & 1019003   &        -                    & 5530718   \\ 
Mn   & 7607   & -        & 23126  & -             & -        & -         &         -                   & 30733     \\ 
Ms   & 5220   & -        & -      & 1631386       & -        & -         &        -                    & 1636606   \\ 
Mt   & -      & -        & -      & -             & -        & 177244    &        -                    & 177244    \\ 
My   & 21497  & -        & 7518   & -             & -        & -         &        -                    & 29015     \\ 
Ro   & 180484 & -        & -      & -             & 610444   & -         &          -                  & 790928    \\ 
Ru   & 208458 & -        & -      & -             & 1640777  & -         &          -                  & 1849235   \\ 
Sr   & 136898 & -        & -      & -             & -        & -         &          -                 & 136898    \\ 
Tr   & 182470 & -        & -      & -             & 205756   & -         &          -                  & 388226    \\ 
Vi   & 171995 & -        & -      & 3055592       & -        & -         &         -                  & 3227587   \\ 
Zh   & 199855 & -        & -      & -             & 25995505 & -         &        -                    & 26195360  \\
Total   & 3245960 & 7442701        & 51724      & 9244442             & 84943811 & 3915046         &    1575608      & 110419292  \\
\bottomrule

\end{tabular}
\caption{Statistics of the dataset \dataset for pre-training. Each entry shows the number of parallel sentence pairs between English and other language X.  }
\label{table-data-description}

\end{table*}

\end{document}